\documentclass{article}
\usepackage[OT1]{fontenc}
\usepackage{iclr2027_conference,times}

\usepackage{graphicx}
\usepackage{amsmath,amssymb}
\usepackage{array,booktabs,tabularx,multirow}
\usepackage{xcolor}
\usepackage{colortbl} 
\usepackage{float}
\usepackage{algorithm}
\usepackage{algpseudocode}
\usepackage[most]{tcolorbox}
\usepackage{listings}
\usepackage{url}
\usepackage{pifont}
\usepackage{placeins}
\usepackage{wrapfig}
\definecolor{EICBlue}{HTML}{54779D}
\definecolor{EICBlueBg}{HTML}{F4F7FB}
\definecolor{EICBlueTitle}{HTML}{E8EFF7}
\definecolor{FindLinkOurs}{HTML}{FDF0E7}
\definecolor{FindLinkBest}{HTML}{EEF8E8}
\definecolor{FindLinkCite}{HTML}{0000AA}
\definecolor{FindLinkGroup}{HTML}{F2F2F2}
\newcommand{\flbest}[1]{\cellcolor{FindLinkBest}#1}
\newcommand{\fltablecite}[1]{{\fontsize{7}{8.5}\selectfont\citep{#1}}}
\newcolumntype{Y}{>{\raggedright\arraybackslash}X}

\newtcblisting{eiccode}[1]{
  enhanced,
  breakable,
  listing only,
  listing engine=listings,
  colback=EICBlueBg,
  colframe=EICBlue!70,
  colbacktitle=EICBlueTitle,
  coltitle=black,
  title={#1},
  fonttitle=\bfseries\small,
  boxrule=0.55pt,
  arc=1.5pt,
  left=5pt,
  right=5pt,
  top=4pt,
  bottom=4pt,
  before skip=6pt,
  after skip=6pt,
  listing options={
    basicstyle=\ttfamily\footnotesize,
    breaklines=true,
    columns=fullflexible,
    keepspaces=true,
    showstringspaces=false,
    literate={²}{{\textsuperscript{2}}}1,
    aboveskip=0pt,
    belowskip=0pt
  }
}

\newtcolorbox{eicnote}[1]{
  enhanced,
  breakable,
  colback=EICBlueBg,
  colframe=EICBlue!70,
  colbacktitle=EICBlueTitle,
  coltitle=black,
  title={#1},
  fonttitle=\bfseries\small,
  boxrule=0.55pt,
  arc=1.5pt,
  left=6pt,
  right=6pt,
  top=5pt,
  bottom=5pt,
  before skip=6pt,
  after skip=6pt
}

\title{FindLink: Learning to Assess Scientific\\Connections from Source-Grounded Premises}

\author{%
\begin{minipage}[t]{\dimexpr\textwidth-2\tabcolsep\relax}
\centering\normalfont\normalsize
{\bfseries
Jingxuan Wei\textsuperscript{1,2,3},\quad
Song Tan\textsuperscript{1},\quad
Xiaohan Liu\textsuperscript{1},\quad
Chao Sang\textsuperscript{1},\quad
Changwei Wang\textsuperscript{2},\\[0.25em]
Linzhuang Sun\textsuperscript{3},\quad
Bihui Yu\textsuperscript{3},\quad
Fuqiang Wang\textsuperscript{1}$^{*}$,\quad
Junnan Zhu\textsuperscript{4}\thanks{Corresponding authors.}\\[0.6em]
}
{\small
\textsuperscript{1}Qilu University of Technology (Shandong Academy of Sciences), Jinan, China\\
\textsuperscript{2}Shandong Computer Science Center, Jinan, China\\
\textsuperscript{3}University of the Chinese Academy of Sciences, Beijing, China\\
\textsuperscript{4}Institute of Automation, Chinese Academy of Sciences, Beijing, China\\[0.6em]
}
{\small\ttfamily
\href{mailto:junnan.zhu@nlpr.ia.ac.cn}{junnan.zhu@nlpr.ia.ac.cn},
\href{mailto:weijingxuan20@mails.ucas.ac.cn}{weijingxuan20@mails.ucas.ac.cn},\\
\href{mailto:10431250277@stu.qlu.edu.cn}{10431250277@stu.qlu.edu.cn}
}
\end{minipage}%
}

\providecommand{\gicmark}{\ding{51}}
\providecommand{\gixmark}{\ding{55}}
\usepackage{hyperref}
\hypersetup{
  colorlinks=true,
  citecolor=FindLinkCite,
  linkcolor=black,
  urlcolor=black
}
\iclrfinalcopy
\hypersetup{
  pdftitle={FindLink: Learning to Assess Scientific Connections from Source-Grounded Premises},
  pdfauthor={Jingxuan Wei, Song Tan, Xiaohan Liu, Chao Sang, Changwei Wang, Linzhuang Sun, Bihui Yu, Fuqiang Wang, Junnan Zhu}
}
\begin{document}

\maketitle
\fancyhead{}
\fancyhead[L]{Preprint}
\renewcommand{\headrulewidth}{0.4pt}
\noindent{\small\textbf{Dataset:} \url{https://huggingface.co/datasets/QLU-LLM/FindLink}}\par
\medskip

\begin{abstract}
Scientific discovery from the literature requires assessing connections among findings obtained under different study conditions. Source attribution makes individual findings traceable, but does not determine what they jointly support. We introduce \textbf{GI-TraceBench}, a benchmark of 10,282 questions in gastrointestinal oncology that unifies provenance question answering, knowledge discovery, and hypothesis evaluation through shared \emph{Paper, Claim, and Evidence Span} records. We further propose \textbf{FindLink}, a two-stage framework for provenance-grounded scientific reasoning. \emph{Premise Grounding} constructs source-bound premises with explicit study context and support status. \emph{Structured Reasoning} organizes these premises into support plans that represent both supported relations and unresolved conditions, then generates evidence-conditioned answers using deterministic compiler feedback. The compiler verifies structural and source binding consistency, but does not assess scientific entailment. On the 1,545-question fixed-evidence test set, FindLink improves the balanced family macro score from 67.95 to 87.25 over direct supervised fine-tuning of the same backbone. A retrained variant without Structured Reasoning achieves slightly higher premise identifier F1, but substantially lower relation and evidence sufficiency F1. These results indicate that scientific reasoning requires not only identifying grounded premises, but also assessing what connections those premises support.

\end{abstract}

\section{Introduction}
\label{sec:introduction}

\begin{figure}[t]
\centering
\includegraphics[
    width=0.88\linewidth]{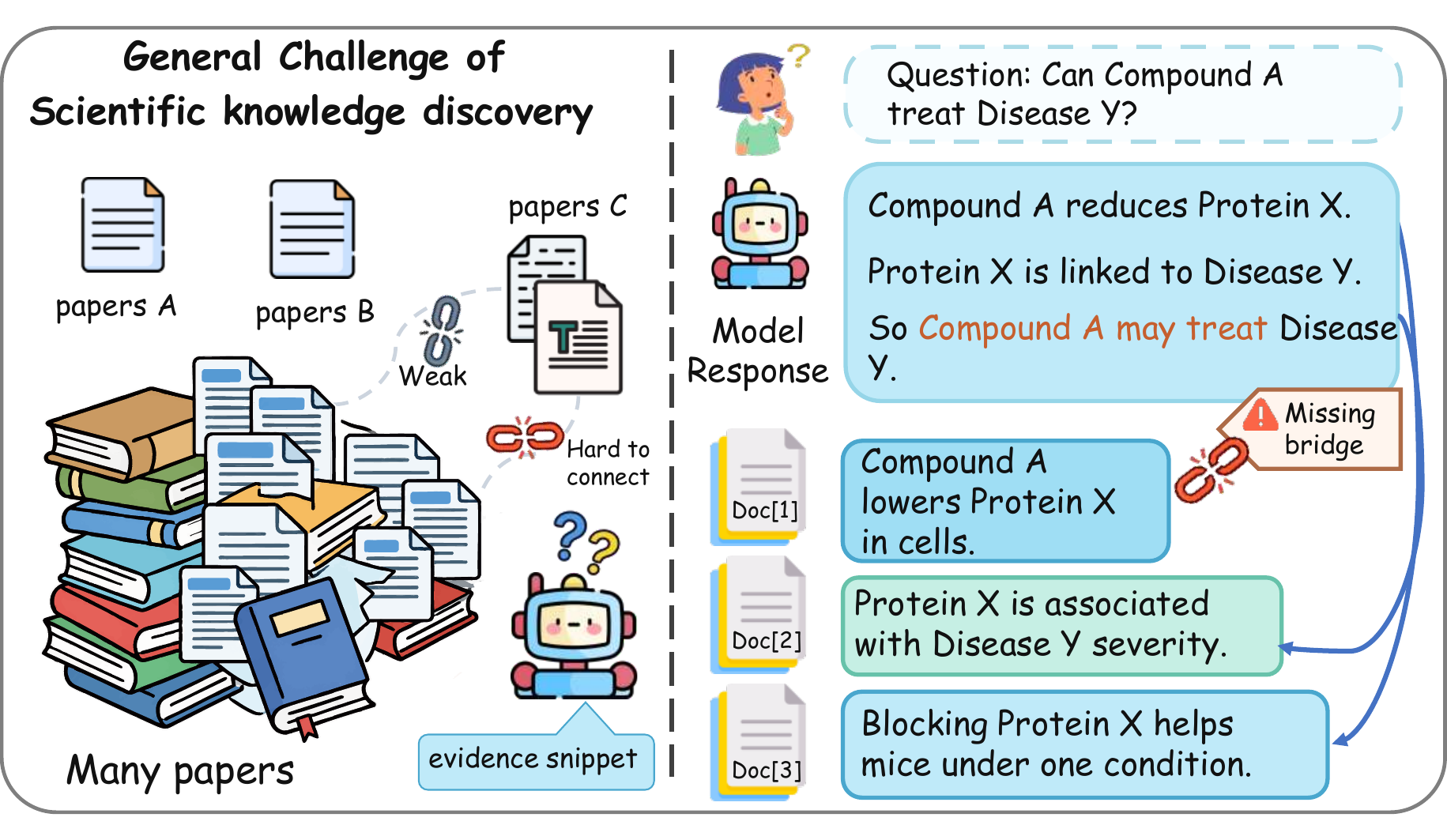}
\caption{\textbf{Overview of Source-Grounded Scientific Connection Assessment.}
Findings from multiple papers may be individually supported but fail to
establish the connecting claim, leaving an evidence gap that provenance-aware
reasoning must identify.}
\label{fig:motivation}
\end{figure}

Scientific knowledge discovery requires connecting findings across studies and determining which connections merit further investigation. Large language models, or LLMs, increasingly support this process through scientific question answering, literature synthesis, and idea generation \citep{asai2026synthesizing,skarlinski2024paperqa2,wang2024scimon}. Citation-aware generation and fine-grained provenance make the sources of individual claims more explicit \citep{gao2023alce,slobodkin2024attribute,wei2026genprove}. However, grounding individual findings does not by itself establish the scientific connection proposed between them. In the schematic example in Figure~\ref{fig:motivation}, a compound lowers a protein in cells, the protein is associated with disease severity, and inhibiting the protein benefits mice under a particular condition. Together, these findings may motivate a treatment hypothesis, but evaluating that hypothesis requires determining whether the intervention, mechanism, and outcome are connected under compatible study conditions, and identifying what evidence is still missing.

We study \emph{provenance-grounded scientific reasoning}, which we define as assessing candidate scientific connections through explicit premises linked to source evidence. This setting requires two complementary capabilities. First, premises must preserve the context in which each finding was established, including study conditions, claim origin, polarity, and support status. They must also distinguish reported observations from author interpretations and proposed explanations. Second, a reasoning system must determine what a set of premises jointly supports, while representing the conditions or relations that remain unresolved. Citation alone cannot establish an unsupported connection, while rejecting all grounded observations because the overall connection is not yet established loses information that remains scientifically useful.

To study these capabilities jointly, we introduce \textbf{GI-TraceBench}, a benchmark containing 10,282 English questions organized into 50 thematic clusters in gastrointestinal oncology. The benchmark comprises three task families, \emph{Provenance QA}, \emph{Knowledge Discovery QA}, and \emph{Hypothesis Evaluation QA}. These tasks respectively evaluate source support, candidate scientific associations, and the evidence and experimental requirements needed to assess proposed hypotheses. All three families share \emph{Paper, Claim, and Evidence Span} records, which connect reasoning outputs to identifiable findings and supporting passages. Cluster disjoint data splits reduce overlap across related scientific topics, while shared cached evidence inputs allow evaluation to focus on how systems interpret supplied evidence rather than on differences in document retrieval.

We propose \textbf{FindLink}, a framework with two stages, \emph{Premise Grounding} and \emph{Structured Reasoning}. Premise Grounding converts evidence into source-bound premises that explicitly encode study context, origin, polarity, and support status. Structured Reasoning then organizes these premises into constrained support plans that represent supported connections together with unresolved relations and study conditions. The final answer is generated from the original evidence, the support plan, and deterministic compiler feedback. Importantly, the compiler verifies representational validity and consistency of source bindings. It does not determine whether a scientific claim is true or whether the evidence entails the proposed connection. Starting from an answer-level supervised fine-tuning checkpoint, FindLink uses sequential adaptation to learn premise construction, support planning, and compiler-conditioned answering.

Across 19 evaluated configurations, FindLink achieves the highest balanced family macro score, improving from 67.95 to 87.25 over direct supervised fine-tuning of the same backbone. The largest improvement occurs on Knowledge Discovery QA, where reasoning depends strongly on assessing relations across multiple grounded findings. A retrained variant without Structured Reasoning achieves slightly higher premise identifier F1, but performs substantially worse on relation, study boundary, and evidence sufficiency F1. This result separates the ability to identify relevant premises from the ability to assess the scientific connections among them.

Our contributions are as follows.
\begin{itemize}
\item We introduce \textbf{GI-TraceBench}, which connects source attribution, scientific connection assessment, and hypothesis evaluation through a shared evidence representation.
\item We introduce \textbf{FindLink}, which learns source-bound premise construction and constrained support planning for evidence-conditioned scientific reasoning.
\item Through fixed evidence evaluation and retrained component variants, we distinguish premise identification from relation, study boundary, and evidence sufficiency assessment.
\end{itemize}

\section{Related Work}
\label{sec:related_work}

\paragraph{Scientific Question Answering and Discovery.}
Scientific QA benchmarks provide complementary evidence settings. PubMedQA and
BioASQ-QA study biomedical questions, while QASPER anchors information-seeking
answers in individual research papers
\citep{jin2019pubmedqa,krithara2023bioasqqa,dasigi2021qasper}. SciFact pairs
scientific claims with support or refutation evidence \citep{wadden2020scifact}.
OpenScholar and PaperQA2 extend scientific answering toward literature synthesis
\citep{asai2026synthesizing,skarlinski2024paperqa2}. Recent benchmarks broaden
the discovery setting: ResearchBench decomposes literature-based
scientific discovery into inspiration retrieval, hypothesis composition,
and pairwise hypothesis ranking \citep{liu2026researchbench}; ScholScan
tests scan-oriented reasoning over complete academic papers
\citep{li2026scholscan}; and MedEvidence evaluates whether models can recover
clinical conclusions from the studies underlying systematic reviews
\citep{polzak2026medevidence}. These settings complement SciMON, ResearchPulse,
and DiscoveryBench, which study inspiration retrieval, method--experiment
chains, and data-driven hypothesis discovery, respectively
\citep{wang2024scimon,chen2025researchpulse,majumder2024discoverybench}.
GI-TraceBench instead evaluates source-grounded premises, study conditions,
candidate relations, unresolved links, and further testing requirements in one
evidence-conditioned benchmark.

\paragraph{Attribution and Fine-Grained Provenance.}
RAG grounds generation in external documents \citep{lewis2020rag}, while ALCE
studies citation-supported answers and citation quality \citep{gao2023alce}.
Attribute First selects evidence before constructing attributable text, and
VeriCite verifies and refines citations
\citep{slobodkin2024attribute,qian2025vericite}. TROVE traces source sentences
and classifies their relationships to target text, while GenProve jointly
generates answers with typed sentence-level provenance
\citep{zhu2025trove,wei2026genprove}. CiteGuard reframes citation validation as
alignment between generated citations and human attribution choices
\citep{choi2026citeguard}; RIGOURATE scores whether claims are proportionate to
their retrieved evidence \citep{james2026rigourate}. These works make citation
or claim-level evidence use more inspectable. FindLink addresses the related
but distinct question of what multiple source-bound premises jointly establish
about a candidate scientific connection. Its intermediate records retain
support status, study scope, polarity, and unresolved links, rather than
treating citation presence or single-claim faithfulness as sufficient for a
new scientific relation.

\paragraph{Structured Reasoning.}
Entailment trees expose intermediate conclusions and supporting facts
\citep{dalvi2021entailment}; Logic-LM and Faithful Chain-of-Thought separate
formal translation from deterministic execution
\citep{pan2023logiclm,lyu2023faithful}; and XGrammar constrains outputs to an
admissible language \citep{dong2024xgrammar}. Recent work makes the missing or
conflicting evidence explicit. Knowing What Is Missing uses an
identify-then-verify procedure for information sufficiency
\citep{jain2026missing}, S2G-RAG maps structured gap judgments to later
retrieval \citep{li2026s2g}, and ConfRAG evaluates reasoning over conflicting
web references \citep{yuan2026confrag}. SCI-Verifier studies cross-disciplinary
scientific answer verification and equivalence judgments
\citep{zheng2026sciverifier}, while CF-RAG uses counterfactual queries to
separate causally decisive evidence from correlated distractors
\citep{qin2026cfrag}. FindLink operates in a different regime: the evidence
packet is fixed, the support plan is restricted, and deterministic compilation
checks representation and source bindings. The compiler does not prove
arbitrary scientific entailment; the final answer may report supported
premises together with an unestablished relation or an unmet testing condition.

\begin{figure}[!htbp]
    \centering
    \includegraphics[
    width=\linewidth
]{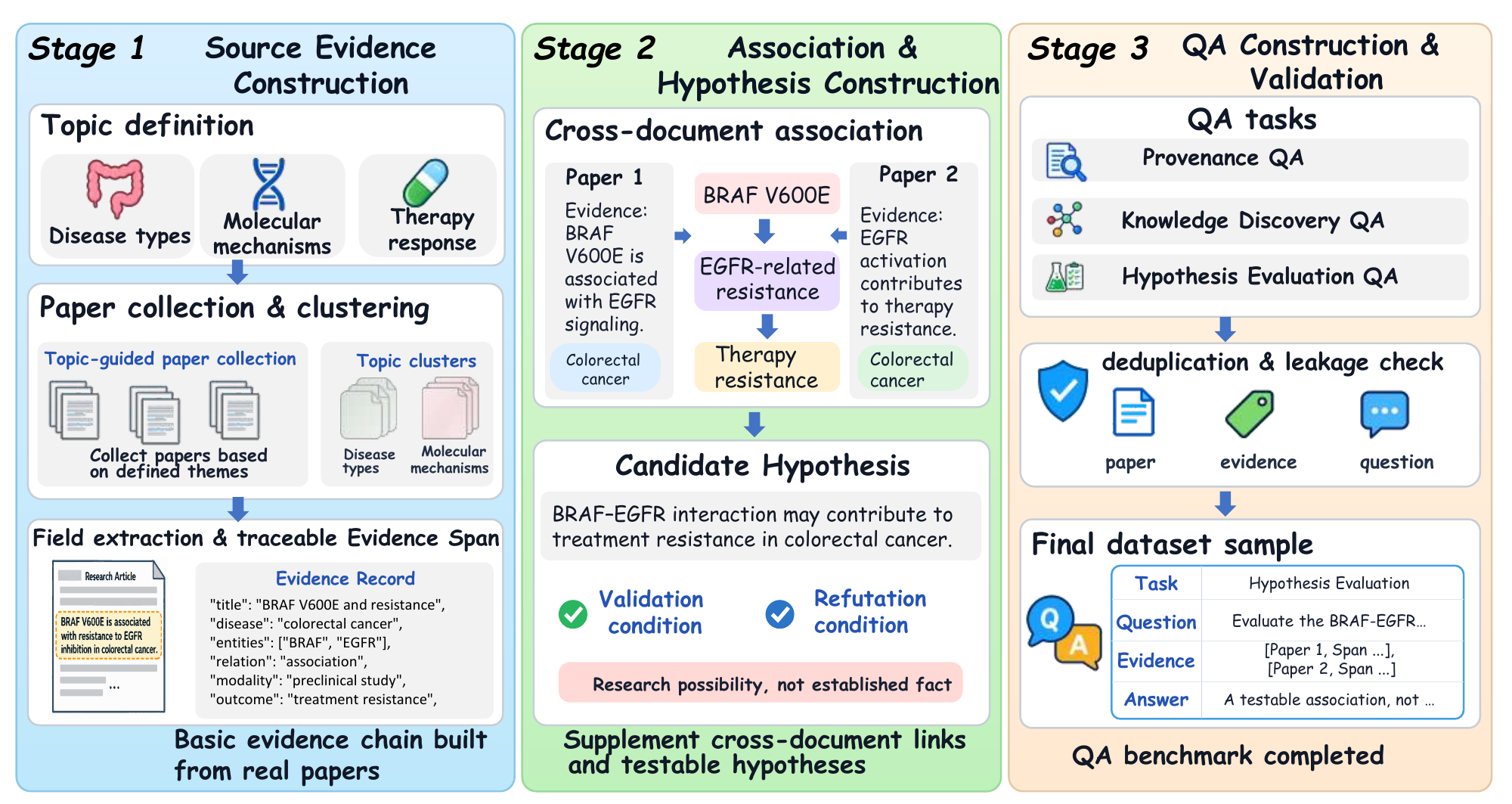}
    \caption{\textbf{GI-TraceBench construction.}
Source papers are converted into linked Paper--Claim--Evidence Span records,
expanded into candidate relations and hypotheses, and transformed into QA
instances after reference, deduplication, and split checks.}
    \label{fig:dataset_construction}
\end{figure}
                                                                            
\section{GI-TraceBench}
\label{sec:dataset}

\subsection{Overview}
\label{sec:dataset_overview}
GI-TraceBench studies three connected aspects of literature-grounded scientific assessment in gastrointestinal oncology.
\emph{Provenance QA} identifies the source passages, support status, and study conditions of a scientific statement.
\emph{Knowledge Discovery QA} assesses a candidate connection through its premise relations, bounded conclusion, and missing evidence.
\emph{Hypothesis Evaluation QA} examines the evidence supporting a candidate hypothesis, the observations needed to test it, and outcomes that would undermine it.
Both single-paper and cross-paper instances are included.

Each instance is $(q,t,E,y^{\mathrm{ref}})$, consisting of a research question, task family, supplied evidence packet, and structured reference answer.
A \emph{Paper} identifies a document; a \emph{Claim} records a statement and contextual qualifications; an \emph{Evidence Span} locates its source passage.
Systems receive $(q,t,E)$, including source text and available public structured records, while $y^{\mathrm{ref}}$ is withheld.
The inputs can therefore include extracted Claims and candidate records in addition to original passages.
The evaluation targets reasoning over this supplied evidence.
Appendix~\ref{app:task_definition} specifies the task-dependent assessment fields.

\subsection{Dataset Construction}
\label{sec:dataset_construction}
\paragraph{Source-linked evidence.}
We organize gastrointestinal oncology literature into thematic clusters and
retain document identifiers, section locations, and available passages,
thereby forming the source-linked evidence stage in
Figure~\ref{fig:dataset_construction}.
Parsing and evidence localization produce Claim records linked to Evidence Spans.
Claims retain contextual qualifications, including polarity and certainty where available; spans preserve the original wording and its location.
Full-text evidence, abstract-only material, and bibliographic pointers are recorded separately.
This stage supplies the source-linked evidence inventory used for candidate
construction and provenance-oriented questions.

\paragraph{Candidate connections and hypotheses.}
Normalized concepts and topic constraints link related Claims into association
records, thereby forming the candidate-construction stage in
Figure~\ref{fig:dataset_construction}.
Each association retains its premise Claims, source references, candidate relation, and contextual boundaries.
Hypothesis cards extend these records with predicted observations, verification requirements, and refutation criteria.
Association-based, typed-chain, and single-paper routes provide complementary constructions.
Deterministic transformations generate these assessment records while retaining candidate status and unresolved evidence gaps.

\paragraph{Question assembly and checks.}
We transform evidence and candidate records into task-specific questions and
structured reference answers, completing the question-assembly and
consistency-checking stage in Figure~\ref{fig:dataset_construction}.
References retain the source bindings and assessment fields needed by each task; integrated provenance questions preserve the individual references of their component checks.
We check reference closure and remove cross-partition conflicts in recorded identifiers and normalized questions.
The supervision includes silver-label components derived from the evidence records.

\begin{wrapfigure}{r}{0.47\textwidth}
    \vspace{-0.5\baselineskip}
    \centering
    \includegraphics[width=\linewidth]
    {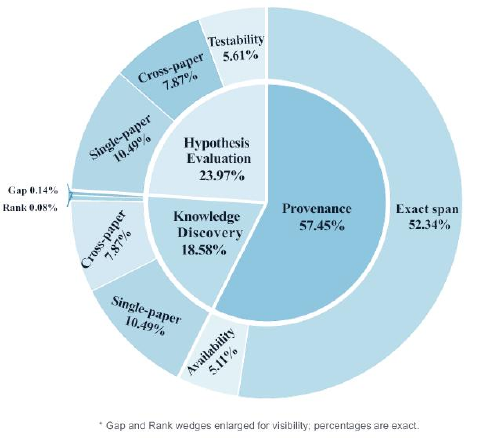}
    \caption{\textbf{GI-TraceBench composition.}
    Task families are shown in the inner disk and their primary
    subtypes in the outer ring.}
    \label{fig:dataset_analysis}
    \vspace{0.3\baselineskip}
\end{wrapfigure}

\subsection{Dataset analysis}
\label{sec:dataset_analysis}

The retained benchmark contains \textbf{10,282} English questions in
\textbf{50} thematic clusters: 5,907 Provenance QA instances (57.4\%),
1,910 Knowledge Discovery QA instances (18.6\%), and 2,465 Hypothesis
Evaluation QA instances (24.0\%). Training uses 42 clusters and 8,737
questions; testing uses eight clusters and 1,545 questions. Figure~\ref{fig:dataset_analysis} summarizes the task-family composition and
primary subtypes of the retained benchmark.

The corpus covers biliary, colorectal, esophageal, gastric, and pancreatic
oncology, together with cross-GI mechanistic and technical themes.
Colorectal and cross-GI questions form the largest shares of the retained QA
instances.
The construction inventory contains 1,614 Paper records, 5,475 Claims,
4,968 Evidence Spans, 733 associations, and 2,048 hypothesis cards.
A source object or hypothesis card can support multiple questions, so
inventory counts and retained QA counts describe different units.
Appendix~\ref{app:dataset_statistics} reports the complete statistics.

\subsection{Comparison with Existing Benchmarks}
\label{sec:benchmark_comparison}
We compare GI-TraceBench with representative scientific QA, provenance, and
discovery benchmarks along three capability axes: evidence grounding,
scientific reasoning, and further testing. As summarized in Appendix~\ref{app:benchmark_comparison} and
Table~\ref{tab:benchmark_comparison}, 
existing resources typically cover only evidence attribution, provenance, or
scientific discovery in isolation
\citep{jin2019pubmedqa,dasigi2021qasper,wei2026genprove,liu2026researchbench}.
GI-TraceBench combines these capabilities through Provenance QA, Knowledge
Discovery QA, and Hypothesis Evaluation QA, and is the only benchmark in the
selected comparison set with explicit coverage of all three axes.

\section{FindLink}
\label{sec:method}

FindLink is an evidence-conditioned framework for scientific assessment.
Given a question, a task family, and a supplied evidence packet, it first
constructs source-bound premises and then organizes them into a constrained
support plan before generating the final answer. The original evidence
remains accessible throughout the pipeline.

Let $x=(q,t,V(E))$ denote the question, task family, and canonical evidence
view. The complete computation is
\begin{equation}
A=G_{\theta_A}(x),\qquad
T=P_{\theta_B}(x,A),\qquad
C=\Gamma_{\mathcal G}(V(E),A,T),\qquad
\hat y=H_{\theta_B}(x,A,T,C).
\label{eq:findlink_pipeline}
\end{equation}
Here, $G_{\theta_A}$ is Premise Grounding, $P_{\theta_B}$ is Structured
Reasoning, $\Gamma_{\mathcal G}$ is a deterministic compiler, and
$H_{\theta_B}$ is compiler-conditioned answer generation.

\begin{figure}[t]
\centering
\includegraphics[width=\linewidth]{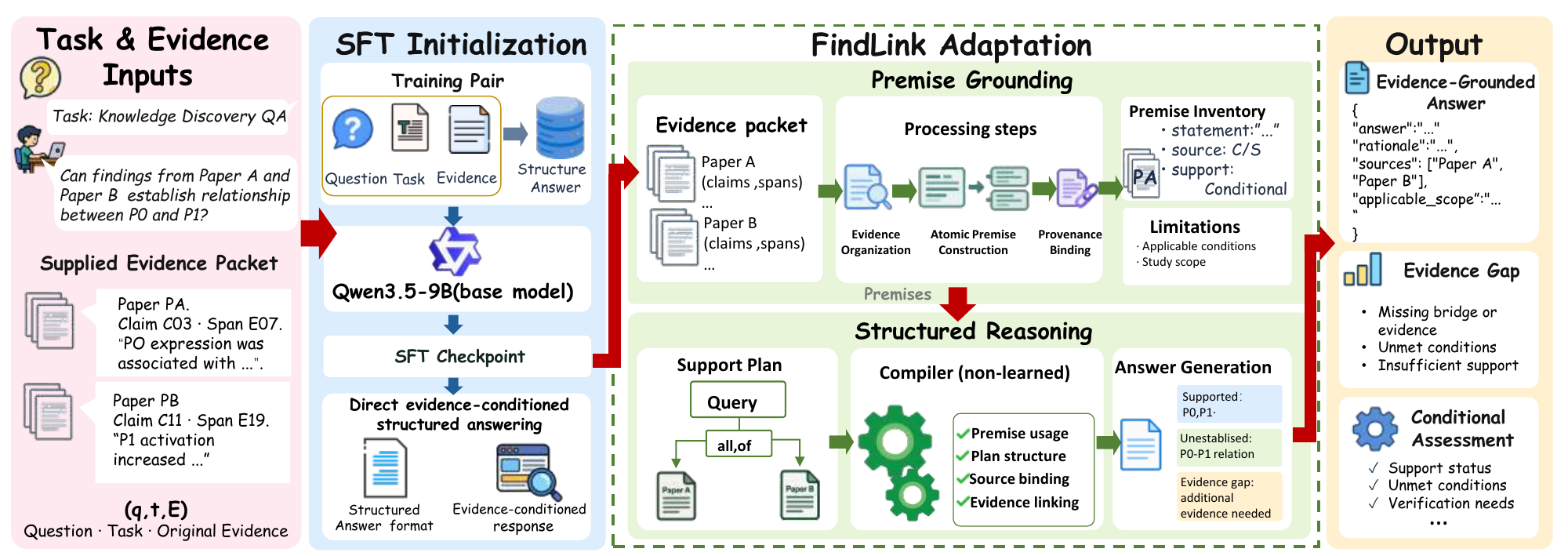}
\caption{\textbf{FindLink framework.}
An answer-level SFT checkpoint is sequentially adapted through Premise
Grounding and Structured Reasoning. The compiler checks the support plan
before generating an evidence-grounded answer with explicit gaps and
conditional assessment fields.}
\label{fig:findlink_architecture}
\end{figure}

Figure~\ref{fig:findlink_architecture} follows this computation. The answer-level SFT checkpoint initializes the pipeline.
FindLink then adds Premise Grounding and Structured Reasoning, while the
compiler checks the intermediate support representation before answer generation.

\subsection{Premise Grounding}
\label{sec:provenance_grounded_reasoning}

Premise Grounding converts the supplied evidence into atomic,
source-bound statements before the model reasons over candidate relations
or hypotheses. The stage returns a premise inventory
$\mathcal P$ and a limitation set $\Lambda$:
\begin{equation}
A=(\mathcal P,\Lambda),
\qquad
p_i=(s_i,c_i,\pi_i,o_i,I_i,\sigma_i).
\label{eq:findlink_ground}
\end{equation}

Each premise records a statement $s_i$, study context $c_i$, polarity
$\pi_i$, origin $o_i$, source pointers $I_i$, and support status
$\sigma_i$. Origins distinguish source observations, author claims, and
candidate statements. An author's proposal therefore supports the
attribution of that proposal, but does not by itself establish the proposed
mechanism.

The inventory is source-bound but not exhaustive: each $I_i$ refers to
canonical evidence pointers, and an empty or incomplete inventory does not
imply that the original evidence is absent. The original evidence remains
available to later support planning through direct source bindings.

\subsection{Structured Reasoning}
\label{sec:structured_reasoning}

Structured Reasoning constructs a constrained support plan rather than an
unrestricted chain of thought. The plan is represented as
\begin{equation}
T=(r,\mathcal N,\mathcal R),
\qquad
T\in\mathcal G(x,A),
\label{eq:findlink_plan}
\end{equation}
where $r$ is the requested target, $\mathcal N$ contains typed statements,
and $\mathcal R$ contains candidate relations, conditions, and validity
judgments. Plan leaves may select generated premises through
\texttt{from\_a}, bind directly to original evidence through
\texttt{from\_source}, or combine existing support expressions through
\texttt{all\_of}. These operators organize the represented assessment and
do not by themselves establish a scientific relation.

The deterministic compiler validates the plan, restores selected evidence,
and records the represented assessment:
\begin{equation}
C=
\Gamma_{\mathcal G}(V(E),A,T)
=
\bigl(\nu_T,\widehat E_T,U_A(T),K_T,\eta_T\bigr).
\label{eq:findlink_compile}
\end{equation}
Here, $\nu_T$ is the structural validity flag, $\widehat E_T$ is the
restored evidence record, $U_A(T)$ is the set of consumed premises,
$K_T$ contains remaining conditions, and $\eta_T$ is the represented
support status.

The compiler checks structure, identifier resolution, and source binding,
but it does not prove scientific entailment. Its feedback conditions the
answer generator:
\begin{equation}
\widetilde y=
\operatorname{Answer}_{\theta_B}(x,A,T,C),
\qquad
\hat y=
\operatorname{Audit}_{\mathcal S_t}(\widetilde y).
\label{eq:findlink_answer}
\end{equation}
The task-specific schema $\mathcal S_t$ requires the answer to preserve
source bindings, study scope, support status, and remaining evidence
requirements. Supported observations may therefore be reported together
with a conditional or unestablished candidate relation.

\subsection{Sequential Supervised Adaptation}
\label{sec:method_training}

We first perform supervised fine-tuning on task-specific,
evidence-conditioned answers:
\begin{equation}
\mathcal L_{\mathrm{SFT}}(\theta)
=
-\sum_i
\log p_\theta
\left(y_i^{\mathrm{ref}}\mid q_i,t_i,E_i\right).
\label{eq:findlink_sft}
\end{equation}
The resulting checkpoint $\theta_{\mathrm{SFT}}$ initializes FindLink.

The two additional stages use different supervision:
\begin{align}
\mathcal L_A(\theta)
&=
-\sum_{(x,A^\ast)\in\mathcal D_A}
\log p_\theta(A^\ast\mid x),\\
\mathcal L_B(\theta)
&=
-\sum_{(x,A,T^\ast)\in\mathcal D_{\mathrm{plan}}}
\log p_\theta(T^\ast\mid x,A) \nonumber\\
&\quad
-\sum_{(x,A,T,C,y^\ast)\in\mathcal D_{\mathrm{answer}}}
\log p_\theta(y^\ast\mid x,A,T,C).
\label{eq:findlink_stage_loss}
\end{align}

The training route is
\begin{equation}
\theta_{\mathrm{SFT}}
\xrightarrow{\mathcal D_A}
\theta_A
\xrightarrow{\mathcal D_{\mathrm{plan}}\cup\mathcal D_{\mathrm{answer}}}
\theta_B.
\label{eq:findlink_training_route}
\end{equation}

The first adaptation stage teaches source-bound premise construction and
limitation recording. The second teaches support-plan construction and
compiler-conditioned answering. Both stages use LoRA adaptation
\citep{hu2021lora}; the compiler remains non-learned. Detailed supervision
records, optimization settings, decoding constraints, and ablations are
provided in Appendix~\ref{app:method_details}.

\section{Experiments}
\label{sec:experiments}

\subsection{Experimental Setup}
\label{sec:experimental_setup}

\paragraph{Data and comparisons.}
All 19 configurations use the complete 1,545-question Test split for
task-specific and text-similarity metrics. For each question, systems
receive the same cached evidence and no reference answer. The comparison includes seven open-weight models, four closed-source models,
four adapted methods (Attribute First~\citep{slobodkin2024attribute},
OpenScholar~\citep{asai2026synthesizing}, PaperQA2~\citep{skarlinski2024paperqa2},
and VeriCite~\citep{qian2025vericite}), Default SFT, FindLink, and two
retrained component variants. Default SFT and the component variants share
FindLink's Qwen3.5-9B backbone~\citep{qwen2026qwen35model}. The adapted methods use the common supplied-evidence interface rather than
their original retrieval pipelines. Default SFT and the two ablation variants
use the same Qwen3.5-9B backbone as FindLink, while backbones, supervision,
and inference budgets vary across the broader comparison.
\paragraph{Training and inference.}
Default SFT uses two LoRA epochs on answer targets from the 8,737-question Train
split. Premise Grounding and Structured Reasoning each add one adaptation epoch.
Greedy decoding uses budgets of 2,048, 2,048, and 4,096 tokens for grounding,
planning, and answering, respectively. Full training and decoding settings are
provided in Appendix~\ref{app:method_details}.

\paragraph{Metrics.}
Provenance Grounding Score (PGS), Discovery Reasoning Score (DRS), and
Hypothesis Structural Score (HSS) assess reference-bound source bindings
and assessment content for the three task families.
Components are computed within applicable clusters and macro-averaged
across clusters; the balanced family macro score (BFM) is the unweighted
mean of PGS, DRS, and HSS.
ROUGE-L, METEOR, BERTScore-F1, and MoverScore measure agreement with
rendered reference answers
\citep{lin2004rouge,banerjee2005meteor,zhang2020bertscore,zhao2019moverscore}.
An LLM judge additionally assigns a holistic 0--5 score across the same
120 questions for all systems, with 40 questions per task family
\citep{findlinkZheng2023Judging}.
This assessment considers answer quality, evidence entailment, citation
correctness and completeness, and boundary control, and is reported
separately from BFM.
Appendix~\ref{app:experimental_protocol} provides the metric definitions,
aggregation rules, and judging criteria.

\begin{table}[t]
\centering
\begingroup
\fontsize{8}{10}\selectfont
\setlength{\tabcolsep}{1.7pt}
\renewcommand{\arraystretch}{1.13}
\setlength{\abovecaptionskip}{5pt}
\setlength{\belowcaptionskip}{5pt}
\hypersetup{citecolor=FindLinkCite,linkcolor=black,urlcolor=black}
\caption{\textbf{Main results and ablations on GI-TraceBench.}
Scores use 0--100, except LLM Judge (0--5).
Bold marks column maxima; green marks the best scores excluding FindLink.
A/B denote Premise Grounding/Structured Reasoning.}
\label{tab:main_results}
\label{tab:ablation_results}
\begin{tabular*}{\linewidth}{
@{\extracolsep{\fill}}l|rrrr|rrrr|c@{}
}
\toprule
Model & PGS$\uparrow$ & DRS$\uparrow$ & HSS$\uparrow$ & BFM$\uparrow$
& R-L$\uparrow$ & MET$\uparrow$ & BERT$\uparrow$ & Mover$\uparrow$
& LLM Judge$\uparrow$ \\
\midrule
\rowcolor{FindLinkGroup}
\multicolumn{10}{@{}c@{}}{
  \textbf{\emph{Open-source models}}
}\\[-1pt]
\midrule
GLM-4-9B-0414~\fltablecite{glm2025glm4model} & 10.29 & 34.94 & 24.01 & 23.08 & 20.11 & 20.67 & 62.94 & 42.93 & 2.75 \\
Gemma-3-12B-IT~\fltablecite{google2025gemma3model} & 12.55 & 21.42 & 24.99 & 19.65 & 16.94 & 24.26 & 54.50 & 36.76 & 2.42 \\
Granite-3.3-8B~\fltablecite{ibm2025granite33model} & 0.97 & 11.91 & 12.76 & 8.55 & 6.46 & 9.78 & 23.82 & 15.84 & 0.91 \\
Hunyuan-7B~\fltablecite{tencent2025hunyuan7b} & 12.41 & 23.18 & 23.46 & 19.68 & 25.18 & 26.42 & 57.78 & 41.12 & 2.17 \\
Llama-3.1-8B-Instruct~\fltablecite{meta2024llama31model} & 11.08 & 8.83 & 1.90 & 7.27 & 5.77 & 6.73 & 14.86 & 10.16 & 0.61 \\
Ministral-3-8B-Instruct-2512~\fltablecite{mistral2025ministral3model} & 4.16 & 11.92 & 10.18 & 8.75 & 11.21 & 15.18 & 29.32 & 20.28 & 1.14 \\
Qwen3.5-9B~\fltablecite{qwen2026qwen35model} & 8.88 & 28.24 & 25.66 & 20.93 & 23.25 & 34.21 & 64.99 & 44.32 & 2.63 \\
\midrule
\rowcolor{FindLinkGroup}
\multicolumn{10}{@{}c@{}}{
  \textbf{\emph{Closed-source models}}
}\\[-1pt]
\midrule
Claude Sonnet 4.6~\fltablecite{anthropic2026sonnet46} & 21.57 & 25.75 & 31.84 & 26.39 & 18.84 & 35.15 & 72.90 & 48.48 & 3.47 \\
Gemini 3.5 Flash~\fltablecite{google2026gemini35flash} & 16.41 & 30.79 & 38.52 & 28.58 & 22.89 & 35.27 & 82.46 & 55.18 & 2.93 \\
GPT-5.4-mini~\fltablecite{openai2026gpt54mini} & 10.28 & 36.32 & 31.81 & 26.14 & 26.89 & 34.74 & 73.33 & 50.90 & 3.60 \\
Qwen3.6-Plus~\fltablecite{alibaba2026qwen36plus} & 10.08 & 38.57 & 39.54 & 29.40 & 30.60 & 37.82 & 80.98 & 57.02 & 3.40 \\
\midrule
\rowcolor{FindLinkGroup}
\multicolumn{10}{@{}c@{}}{
  \textbf{\emph{Adapted methods}}
}\\[-1pt]
\midrule
Attribute First~\fltablecite{slobodkin2024attribute} & 9.12 & 35.29 & 36.68 & 27.03 & 20.40 & 26.18 & 70.37 & 47.43 & 1.89 \\
OpenScholar~\fltablecite{asai2026synthesizing} & 7.67 & 39.23 & 42.06 & 29.65 & 22.00 & 33.23 & 83.03 & 56.04 & 2.10 \\
PaperQA2~\fltablecite{skarlinski2024paperqa2} & 9.38 & 34.91 & 40.34 & 28.21 & 10.85 & 21.04 & 78.57 & 52.36 & 3.75 \\
VeriCite~\fltablecite{qian2025vericite} & 10.05 & 34.73 & 38.66 & 27.81 & 25.60 & 34.61 & 82.52 & 56.95 & 2.52 \\
\midrule
\rowcolor{FindLinkGroup}
\multicolumn{10}{@{}c@{}}{
  \textbf{\emph{Ablation analysis}}
}\\[-1pt]
\midrule
Default SFT & 72.72 & 60.80 & 70.32 & 67.95 & 52.89 & 76.26 & 80.65 & 65.36 & 3.68 \\
w/o A & 73.17 & \flbest{83.74} & 87.22 & 81.38 & 58.01 & 77.88 & 89.24 & 70.03 & 3.89 \\
w/o B & \flbest{79.55} & 79.22 & \flbest{\textbf{91.93}} & \flbest{83.57} & \flbest{60.95} & \flbest{83.31} & \flbest{\textbf{94.18}} & \flbest{73.36} & \flbest{3.90} \\
\midrule
\rowcolor{FindLinkOurs}
\textbf{FindLink (Ours)} & \textbf{84.03} & \textbf{87.94} & 89.79 & \textbf{87.25} & \textbf{61.81} & \textbf{84.40} & 93.04 & \textbf{73.67} & \textbf{4.28} \\
\bottomrule
\end{tabular*}
\endgroup
\end{table}

\subsection{Main Results}
\label{sec:main_results}

FindLink obtains the highest BFM in Table~\ref{tab:main_results}, reaching 87.25.
Relative to Default SFT, PGS increases by 11.31 points, DRS by 27.14 points, and HSS by 19.47 points, for a 19.30-point BFM improvement.
The largest gain occurs in Knowledge Discovery QA, whose assessment includes premise relations, study boundaries, evidence sufficiency, and missing evidence.
The same-backbone comparison provides the primary reference for evaluating the complete adaptation and inference pipeline.

\paragraph{Answer quality.}
\label{sec:llm_assessment}
FindLink improves all four text-similarity scores over Default SFT, with BERTScore-F1 increasing from 80.65 to 93.04.
It also receives the highest LLM-judge score, 4.28, compared with 3.68 for Default SFT and 3.75 for PaperQA2, the strongest adapted method on this measure.
The higher text-similarity scores coincide with higher task-family scores.
These comparisons use the evaluation scopes specified above: the full Test split for automatic metrics and the shared 120-question subset for the judge.

\subsection{Ablation Study}
\label{sec:ablation_study}

The bottom block of Table~\ref{tab:main_results} compares FindLink with two
retrained variants. Removing Premise Grounding (\emph{w/o A}) lowers PGS from
84.03 to 73.17 and BFM from 87.25 to 81.38, indicating weaker premise
grounding and overall task performance. Removing Structured Reasoning
(\emph{w/o B}) improves HSS from 89.79 to 91.93 and BERTScore-F1 from 93.04
to 94.18, but lowers DRS from 87.94 to 79.22, BFM from 87.25 to 83.57, and
the LLM-Judge score from 4.28 to 3.90. Additional training and interface
details are provided in Appendix~\ref{app:ablation_details}. These results
suggest that direct answering can improve surface-level similarity on some
tasks, whereas structured reasoning is needed for reliable discovery and
balanced performance. The full FindLink pipeline therefore provides the best
overall trade-off across grounding, reasoning, and answer quality.

\begin{figure}[t]
    \centering
    \includegraphics[
        width=\linewidth,
        trim={0 18pt 0 0},
        clip
    ]{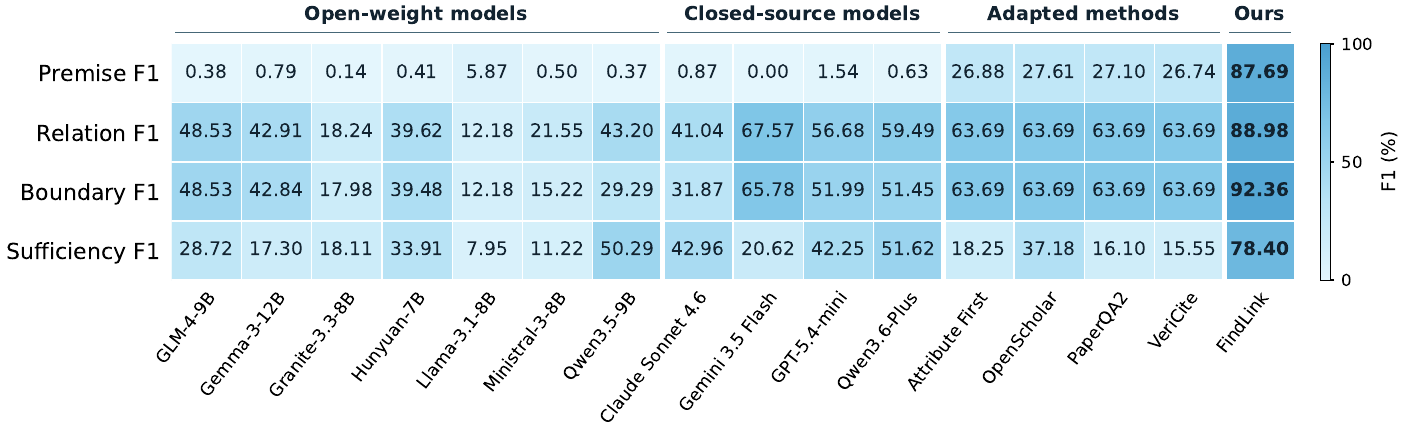}
    \caption{\textbf{Fine-grained knowledge-discovery performance.}
    Columns show seven open-weight models, four closed-source models, four
    adapted methods, and FindLink; rows show the four F1 components of DRS.
    The shared color scale is 0--100. Default SFT and ablations remain in
    Table~\ref{tab:discovery_f1_all}.}
    \label{fig:diag_discovery_f1}
\end{figure}

\subsection{Diagnostic Analysis}
\label{sec:diagnostic_analysis}

\paragraph{Fine-grained discovery performance.}
Figure~\ref{fig:diag_discovery_f1} compares the four discovery F1 components
across 16 systems. FindLink leads the plotted systems in premise,
relation, boundary, and sufficiency F1 (87.69, 88.98, 92.36, and 78.40).
Sufficiency remains the lowest-scoring component; it distinguishes
answerable, conservative, and abstention outcomes. This motivates
examining whether low scores reflect missing evidence or an incorrect
judgment about evidence that is already available. The full component
results and their extraction rules are given in
Appendix~\ref{app:discovery_f1_all}.

\paragraph{Qualitative diagnostics.}
Qualitative cases complement the component analysis by illustrating
candidate--verification misalignment, study-context boundary violations,
bounded assessment, and provenance localization. The corresponding cases and
figures are provided in Appendix~\ref{app:qualitative_cases}.

\subsection{Consistency with Human Evaluation}
\label{sec:human_llm_correlation}

\begin{wrapfigure}{r}{0.47\textwidth}
    \vspace{-0.5\baselineskip}
    \centering
    \includegraphics[width=\linewidth]{
        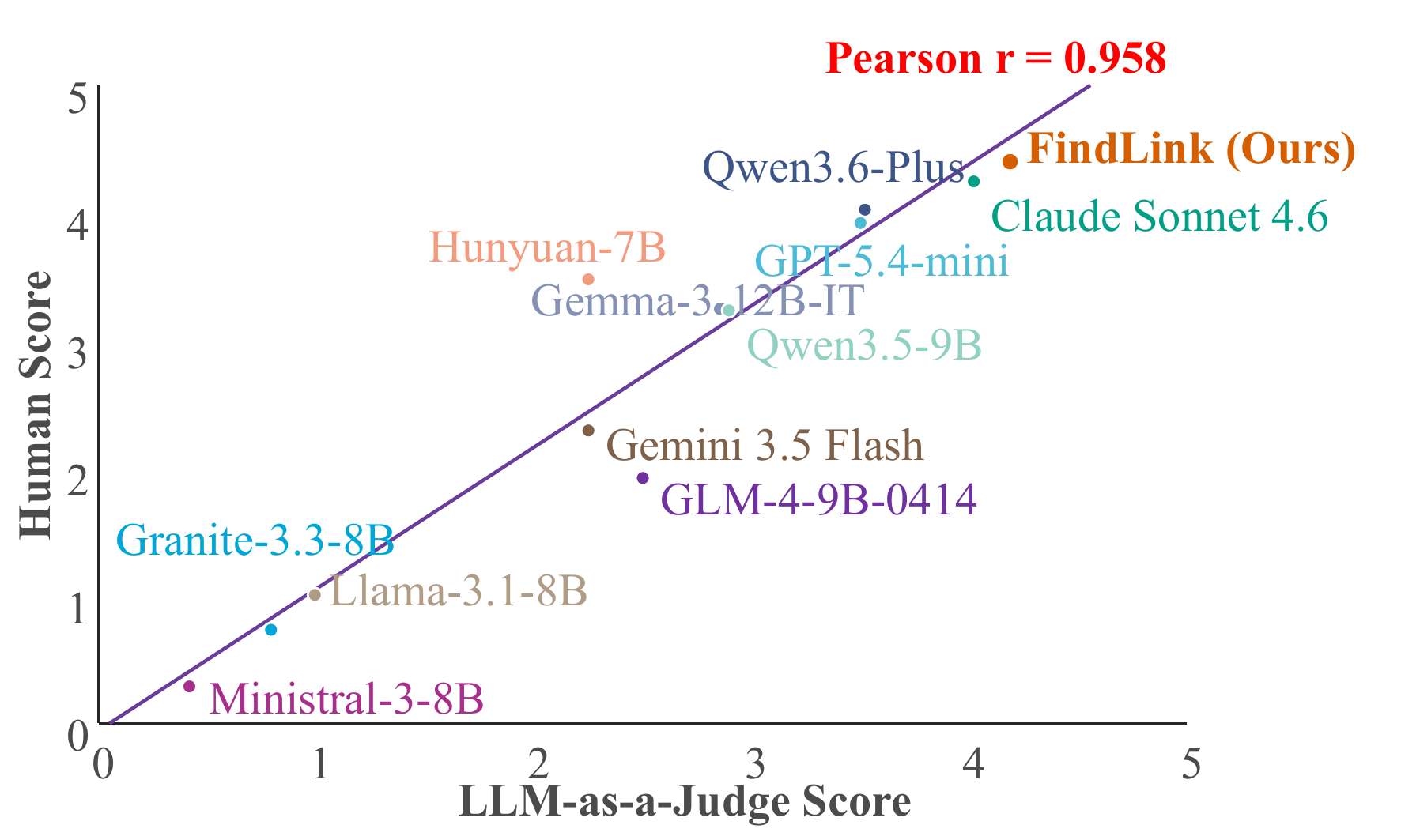}
    \caption{\textbf{Consistency between LLM-as-Judge and human evaluation.}}
    \label{fig:human_llm_correlation}
    \vspace{0.3\baselineskip}
\end{wrapfigure}

To assess the reliability of LLM-as-a-Judge as a model-level signal for
evidence-grounded answer quality, we measure its consistency with human
ratings on a shared, pre-specified 24-question subset. Figure~\ref{fig:human_llm_correlation}
shows strong system-level agreement across 12 systems, including 11 open- and
closed-source comparison models and FindLink, with a Pearson correlation of
$r=0.958$. The alignment spans the full score range, from low-scoring
open-weight baselines to higher-scoring closed-source models and FindLink.
This broad coverage suggests that the judge captures meaningful system-level
differences rather than merely separating a small cluster of similarly
performing models. Detailed scoring protocols, aligned system means, and
reviewer-agreement analyses are provided in Appendices~\ref{app:human_llm_correlation}
and \ref{app:human_agreement_results}.

\section{Conclusion}
\label{sec:conclusion}

We introduce GI-TraceBench and FindLink for assessing scientific connections from source-grounded premises. GI-TraceBench unifies provenance, knowledge discovery, and hypothesis evaluation through shared evidence records, while FindLink learns premises and structured support plans that preserve study context and unresolved conditions. Under the fixed evidence evaluation, FindLink improves balanced performance over direct supervised fine-tuning, and component analysis separates premise identification from relation, study boundary, and evidence sufficiency assessment. These results suggest that provenance-grounded scientific reasoning should preserve both supported observations and the requirements of connections that remain unestablished, rather than treating citation presence or structural validity as scientific support. Our current evaluation is limited to a single biomedical domain and reference-bound evidence. Future work should examine cross-domain generalization and stronger methods for premise verification.

\label{sec:main_text_end}

\section*{AI Use Statement}
We used generative AI tools to assist us in processing the literature and
constructing the dataset. We reviewed and verified the resulting records and
labels through the documented audit procedure, and we take responsibility for
the final dataset, experiments, and claims.

\section*{Reproducibility Statement}
Appendix~\ref{app:dataset_details} documents the evidence representation, construction process, split identifiers, and dataset statistics.
Appendix~\ref{app:method_details} describes supervision, training settings, decoding, and component interfaces.
Appendices~\ref{app:experimental_protocol} and~\ref{app:prompt_specifications} specify the evaluation protocol and prompt interfaces.
The reported numerical comparisons use the fixed evidence and evaluation subsets described in these sections.

\FloatBarrier
\bibliography{iclr2027_conference}

\begin{thebibliography}{48}
\providecommand{\natexlab}[1]{#1}
\providecommand{\url}[1]{\texttt{#1}}
\expandafter\ifx\csname urlstyle\endcsname\relax
  \providecommand{\doi}[1]{doi: #1}\else
  \providecommand{\doi}{doi: \begingroup \urlstyle{rm}\Url}\fi

\bibitem[{Alibaba Cloud}(2026)]{alibaba2026qwen36plus}
{Alibaba Cloud}.
\newblock {Qwen3.6-Plus: Official Model Documentation}.
\newblock Alibaba Cloud Model Studio, 2026.
\newblock URL
  \url{https://docs.modelstudio.console.alibabacloud.com/en/model-studio/qwen3-6-plus}.
\newblock Accessed 19 September 2026.

\bibitem[{Anthropic}(2026)]{anthropic2026sonnet46}
{Anthropic}.
\newblock {Introducing Sonnet 4.6}.
\newblock Official release announcement, 2026.
\newblock URL \url{https://www.anthropic.com/news/claude-sonnet-4-6}.
\newblock Accessed 19 September 2026.

\bibitem[Asai et~al.(2026)Asai, He, Shao, Shi, Singh, Chang, Lo, Soldaini,
  Feldman, D'arcy, et~al.]{asai2026synthesizing}
Akari Asai, Jacqueline He, Rulin Shao, Weijia Shi, Amanpreet Singh, Joseph~Chee
  Chang, Kyle Lo, Luca Soldaini, Sergey Feldman, Mike D'arcy, et~al.
\newblock Synthesizing scientific literature with retrieval-augmented language
  models.
\newblock \emph{Nature}, 650\penalty0 (8103):\penalty0 857--863, 2026.

\bibitem[Banerjee \& Lavie(2005)Banerjee and Lavie]{banerjee2005meteor}
Satanjeev Banerjee and Alon Lavie.
\newblock Meteor: An automatic metric for mt evaluation with improved
  correlation with human judgments.
\newblock In \emph{Proceedings of the acl workshop on intrinsic and extrinsic
  evaluation measures for machine translation and/or summarization}, pp.\
  65--72, 2005.

\bibitem[Chen et~al.(2025)Chen, Wei, Yao, Wang, Wu, Yu, Li, and
  Tan]{chen2025researchpulse}
Qi~Chen, Jingxuan Wei, Zhuoya Yao, Haiguang Wang, Gaowei Wu, Bihui Yu, Siyuan
  Li, and Cheng Tan.
\newblock Researchpulse: Building method-experiment chains through
  multi-document scientific inference.
\newblock In \emph{Proceedings of the 33rd ACM International Conference on
  Multimedia}, pp.\  9110--9119, 2025.

\bibitem[Choi et~al.(2026)Choi, Guo, Fung, and Wang]{choi2026citeguard}
Yee~Man Choi, Xuehang Guo, Yi~R Fung, and Qingyun Wang.
\newblock Citeguard: Faithful citation attribution for llms via
  retrieval-augmented validation.
\newblock In \emph{Proceedings of the 64th Annual Meeting of the Association
  for Computational Linguistics (Volume 1: Long Papers)}, pp.\  6241--6257,
  2026.

\bibitem[Dalvi et~al.(2021)Dalvi, Jansen, Tafjord, Xie, Smith, Pipatanangkura,
  and Clark]{dalvi2021entailment}
Bhavana Dalvi, Peter Jansen, Oyvind Tafjord, Zhengnan Xie, Hannah Smith,
  Leighanna Pipatanangkura, and Peter Clark.
\newblock Explaining answers with entailment trees.
\newblock In \emph{Proceedings of the 2021 conference on empirical methods in
  natural language processing}, pp.\  7358--7370, 2021.

\bibitem[Dasigi et~al.(2021)Dasigi, Lo, Beltagy, Cohan, Smith, and
  Gardner]{dasigi2021qasper}
Pradeep Dasigi, Kyle Lo, Iz~Beltagy, Arman Cohan, Noah~A Smith, and Matt
  Gardner.
\newblock A dataset of information-seeking questions and answers anchored in
  research papers.
\newblock In \emph{Proceedings of the 2021 Conference of the North American
  Chapter of the Association for Computational Linguistics: Human Language
  Technologies}, pp.\  4599--4610, 2021.

\bibitem[Dong et~al.(2025)Dong, Ruan, Cai, Xu, Zhao, Lai, and
  Chen]{dong2024xgrammar}
Yixin Dong, Charlie~F Ruan, Yaxing Cai, Ziyi Xu, Yilong Zhao, Ruihang Lai, and
  Tianqi Chen.
\newblock Xgrammar: Flexible and efficient structured generation engine for
  large language models.
\newblock \emph{Proceedings of Machine Learning and Systems}, 7, 2025.

\bibitem[Gao et~al.(2023)Gao, Yen, Yu, and Chen]{gao2023alce}
Tianyu Gao, Howard Yen, Jiatong Yu, and Danqi Chen.
\newblock Enabling large language models to generate text with citations.
\newblock In \emph{Proceedings of the 2023 Conference on Empirical Methods in
  Natural Language Processing}, pp.\  6465--6488, 2023.

\bibitem[{Google DeepMind}(2025)]{google2025gemma3model}
{Google DeepMind}.
\newblock {Gemma-3-12B-IT: Official Model Card}.
\newblock Hugging Face model repository, 2025.
\newblock URL \url{https://huggingface.co/google/gemma-3-12b-it}.
\newblock Accessed 19 September 2026.

\bibitem[{Google DeepMind}(2026)]{google2026gemini35flash}
{Google DeepMind}.
\newblock {Gemini 3.5 Flash: Model Card}.
\newblock Official model card, 2026.
\newblock URL
  \url{https://deepmind.google/models/model-cards/gemini-3-5-flash/}.
\newblock Accessed 19 September 2026.

\bibitem[Hu et~al.(2021)Hu, Shen, Wallis, Allen-Zhu, Li, Wang, Wang, and
  Chen]{hu2021lora}
Edward~J Hu, Yelong Shen, Phillip Wallis, Zeyuan Allen-Zhu, Yuanzhi Li, Shean
  Wang, Lu~Wang, and Weizhu Chen.
\newblock Lora: Low-rank adaptation of large language models.
\newblock \emph{arXiv preprint arXiv:2106.09685}, 2021.

\bibitem[{IBM}(2025)]{ibm2025granite33model}
{IBM}.
\newblock {Granite-3.3-8B-Instruct: Official Model Card}.
\newblock Hugging Face model repository, 2025.
\newblock URL \url{https://huggingface.co/ibm-granite/granite-3.3-8b-instruct}.
\newblock Accessed 19 September 2026.

\bibitem[Jain \& Garimella(2026)Jain and Garimella]{jain2026missing}
Akriti Jain and Aparna Garimella.
\newblock Knowing what's missing: Assessing information sufficiency in question
  answering.
\newblock In \emph{Findings of the Association for Computational Linguistics:
  EACL 2026}, pp.\  4163--4174, 2026.

\bibitem[James et~al.(2026)James, Xiao, Li, Moosavi, and
  Lin]{james2026rigourate}
Joseph James, Chenghao Xiao, Yucheng Li, Nafise~Sadat Moosavi, and Chenghua
  Lin.
\newblock Rigourate: Quantifying scientific exaggeration with evidence-aligned
  claim evaluation.
\newblock In \emph{Findings of the Association for Computational Linguistics:
  ACL 2026}, pp.\  34022--34043, 2026.

\bibitem[Jin et~al.(2019)Jin, Dhingra, Liu, Cohen, and Lu]{jin2019pubmedqa}
Qiao Jin, Bhuwan Dhingra, Zhengping Liu, William Cohen, and Xinghua Lu.
\newblock Pubmedqa: A dataset for biomedical research question answering.
\newblock In \emph{Proceedings of the 2019 conference on empirical methods in
  natural language processing and the 9th international joint conference on
  natural language processing (EMNLP-IJCNLP)}, pp.\  2567--2577, 2019.

\bibitem[Krithara et~al.(2023)Krithara, Nentidis, Bougiatiotis, and
  Paliouras]{krithara2023bioasqqa}
Anastasia Krithara, Anastasios Nentidis, Konstantinos Bougiatiotis, and
  Georgios Paliouras.
\newblock Bioasq-qa: A manually curated corpus for biomedical question
  answering.
\newblock \emph{Scientific data}, 10\penalty0 (1):\penalty0 170, 2023.

\bibitem[Lewis et~al.(2020)Lewis, Perez, Piktus, Petroni, Karpukhin, Goyal,
  K{\"u}ttler, Lewis, Yih, Rockt{\"a}schel, et~al.]{lewis2020rag}
Patrick Lewis, Ethan Perez, Aleksandra Piktus, Fabio Petroni, Vladimir
  Karpukhin, Naman Goyal, Heinrich K{\"u}ttler, Mike Lewis, Wen-tau Yih, Tim
  Rockt{\"a}schel, et~al.
\newblock Retrieval-augmented generation for knowledge-intensive nlp tasks.
\newblock \emph{Advances in neural information processing systems},
  33:\penalty0 9459--9474, 2020.

\bibitem[Li et~al.(2026{\natexlab{a}})Li, Zou, Lv, Zhang, and Zhou]{li2026s2g}
Minghan Li, Junjie Zou, Xinxuan Lv, Chao Zhang, and Guodong Zhou.
\newblock S2g-rag: Structured sufficiency and gap judging for iterative
  retrieval-augmented qa.
\newblock In \emph{Proceedings of the 64th Annual Meeting of the Association
  for Computational Linguistics (Volume 1: Long Papers)}, pp.\  25846--25862,
  2026{\natexlab{a}}.

\bibitem[Li et~al.(2026{\natexlab{b}})Li, Tang, Wang, Hu, Wang, Lu, Huang,
  Chen, Tan, Liu, et~al.]{li2026scholscan}
Rongjin Li, Zichen Tang, Xianghe Wang, Xinyi Hu, Zhengyu Wang, Zhengyu Lu,
  Yiling Huang, Jiayuan Chen, Weisheng Tan, Jiacheng Liu, et~al.
\newblock Not search, but scan: Benchmarking mllms on scan-oriented academic
  paper reasoning.
\newblock In \emph{The Fourteenth International Conference on Learning
  Representations}, 2026{\natexlab{b}}.

\bibitem[Lin(2004)]{lin2004rouge}
Chin-Yew Lin.
\newblock Rouge: A package for automatic evaluation of summaries.
\newblock In \emph{Text summarization branches out}, pp.\  74--81, 2004.

\bibitem[Liu et~al.(2019)Liu, Ott, Goyal, Du, Joshi, Chen, Levy, Lewis,
  Zettlemoyer, and Stoyanov]{liu2019roberta}
Yinhan Liu, Myle Ott, Naman Goyal, Jingfei Du, Mandar Joshi, Danqi Chen, Omer
  Levy, Mike Lewis, Luke Zettlemoyer, and Veselin Stoyanov.
\newblock Roberta: A robustly optimized bert pretraining approach.
\newblock \emph{arXiv preprint arXiv:1907.11692}, 2019.

\bibitem[Liu et~al.(2026)Liu, Yang, Xie, Ni, Gao, Li, Tang, Ouyang, Cambria,
  and Zhou]{liu2026researchbench}
Yujie Liu, Zonglin Yang, Tong Xie, Jinjie Ni, Ben Gao, Yuqiang Li, Shixiang
  Tang, Wanli Ouyang, Erik Cambria, and Dongzhan Zhou.
\newblock Researchbench: Benchmarking llms in scientific discovery via
  inspiration-based task decomposition.
\newblock In \emph{Findings of the Association for Computational Linguistics:
  ACL 2026}, pp.\  13187--13207, 2026.

\bibitem[Lyu et~al.(2023)Lyu, Havaldar, Stein, Zhang, Rao, Wong, Apidianaki,
  and Callison-Burch]{lyu2023faithful}
Qing Lyu, Shreya Havaldar, Adam Stein, Li~Zhang, Delip Rao, Eric Wong, Marianna
  Apidianaki, and Chris Callison-Burch.
\newblock Faithful chain-of-thought reasoning.
\newblock In \emph{Proceedings of the 13th international joint conference on
  natural language processing and the 3rd conference of the Asia-Pacific
  Chapter of the Association for Computational Linguistics (Volume 1: Long
  papers)}, pp.\  305--329, 2023.

\bibitem[Majumder et~al.(2025)Majumder, Surana, Agarwal, Dalvi~Mishra, Meena,
  Prakhar, Vora, Khot, Sabharwal, and Clark]{majumder2024discoverybench}
Bodhisattwa~Prasad Majumder, Harshit Surana, Dhruv Agarwal, Bhavana
  Dalvi~Mishra, Abhijeetsingh Meena, Aryan Prakhar, Tirth Vora, Tushar Khot,
  Ashish Sabharwal, and Peter Clark.
\newblock Discoverybench: Towards data-driven discovery with large language
  models.
\newblock In \emph{International Conference on Learning Representations},
  volume 2025, pp.\  4556--4579, 2025.

\bibitem[{Meta}(2024)]{meta2024llama31model}
{Meta}.
\newblock {Llama-3.1-8B-Instruct: Official Model Card}.
\newblock Hugging Face model repository, 2024.
\newblock URL \url{https://huggingface.co/meta-llama/Llama-3.1-8B-Instruct}.
\newblock Accessed 19 September 2026.

\bibitem[{Mistral AI}(2025)]{mistral2025ministral3model}
{Mistral AI}.
\newblock {Ministral-3-8B-Instruct-2512: Official Model Card}.
\newblock Hugging Face model repository, 2025.
\newblock URL
  \url{https://huggingface.co/mistralai/Ministral-3-8B-Instruct-2512}.
\newblock Accessed 19 September 2026.

\bibitem[{OpenAI}(2026)]{openai2026gpt54mini}
{OpenAI}.
\newblock {Introducing GPT-5.4 mini and nano}.
\newblock Official release announcement, 2026.
\newblock URL
  \url{https://openai.com/index/introducing-gpt-5-4-mini-and-nano/}.
\newblock Accessed 19 September 2026.

\bibitem[Pan et~al.(2023)Pan, Albalak, Wang, and Wang]{pan2023logiclm}
Liangming Pan, Alon Albalak, Xinyi Wang, and William Wang.
\newblock Logic-lm: Empowering large language models with symbolic solvers for
  faithful logical reasoning.
\newblock In \emph{Findings of the Association for Computational Linguistics:
  EMNLP 2023}, pp.\  3806--3824, 2023.

\bibitem[Polzak et~al.(2026)Polzak, Lozano, Sun, Burgess, Zhang, Wu, Chiang,
  Nirschl, and Yeung]{polzak2026medevidence}
Christopher Polzak, Alejandro Lozano, Min Sun, James Burgess, Yuhui Zhang,
  Kevin Wu, Chia-Chun Chiang, Jeffrey~J Nirschl, and Serena Yeung.
\newblock Can large language models match the conclusions of systematic
  reviews?
\newblock In \emph{International Conference on Learning Representations},
  volume 2026, pp.\  34089--34102, 2026.

\bibitem[Qian et~al.(2025)Qian, Fan, Guo, Zhang, Chen, Yin, and
  Cheng]{qian2025vericite}
Haosheng Qian, Yixing Fan, Jiafeng Guo, Ruqing Zhang, Qi~Chen, Dawei Yin, and
  Xueqi Cheng.
\newblock Vericite: Towards reliable citations in retrieval-augmented
  generation via rigorous verification.
\newblock In \emph{Proceedings of the 2025 Annual International ACM SIGIR
  Conference on Research and Development in Information Retrieval in the Asia
  Pacific Region}, pp.\  47--54, 2025.

\bibitem[Qin et~al.(2026)Qin, Wei, Chen, and Wang]{qin2026cfrag}
Huaiyu Qin, Chunyu Wei, Yueguo Chen, and Yunhai Wang.
\newblock Counterfactual reasoning for retrieval-augmented generation.
\newblock In \emph{International Conference on Learning Representations},
  volume 2026, pp.\  16683--16714, 2026.

\bibitem[{Qwen Team}(2026)]{qwen2026qwen35model}
{Qwen Team}.
\newblock {Qwen3.5-9B}: Official model card.
\newblock Hugging Face model repository, 2026.
\newblock URL \url{https://huggingface.co/Qwen/Qwen3.5-9B}.
\newblock Accessed 19 September 2026.

\bibitem[Skarlinski et~al.(2024)Skarlinski, Cox, Laurent, Braza, Hinks,
  Hammerling, Ponnapati, Rodriques, and White]{skarlinski2024paperqa2}
Michael~D Skarlinski, Sam Cox, Jon~M Laurent, James~D Braza, Michaela Hinks,
  Michael~J Hammerling, Manvitha Ponnapati, Samuel~G Rodriques, and Andrew~D
  White.
\newblock Language agents achieve superhuman synthesis of scientific knowledge.
\newblock \emph{arXiv preprint arXiv:2409.13740}, 2024.

\bibitem[Slobodkin et~al.(2024)Slobodkin, Hirsch, Cattan, Schuster, and
  Dagan]{slobodkin2024attribute}
Aviv Slobodkin, Eran Hirsch, Arie Cattan, Tal Schuster, and Ido Dagan.
\newblock Attribute first, then generate: Locally-attributable grounded text
  generation.
\newblock In \emph{Proceedings of the 62nd Annual Meeting of the Association
  for Computational Linguistics (Volume 1: Long Papers)}, pp.\  3309--3344,
  2024.

\bibitem[{Tencent Hunyuan Team}(2025)]{tencent2025hunyuan7b}
{Tencent Hunyuan Team}.
\newblock {Hunyuan-7B-Instruct: Official Model Card}.
\newblock Hugging Face model repository, 2025.
\newblock URL \url{https://huggingface.co/tencent/Hunyuan-7B-Instruct}.
\newblock Accessed 19 September 2026.

\bibitem[Wadden et~al.(2020)Wadden, Lin, Lo, Wang, van Zuylen, Cohan, and
  Hajishirzi]{wadden2020scifact}
David Wadden, Shanchuan Lin, Kyle Lo, Lucy~Lu Wang, Madeleine van Zuylen, Arman
  Cohan, and Hannaneh Hajishirzi.
\newblock Fact or fiction: Verifying scientific claims.
\newblock In \emph{Proceedings of the 2020 conference on empirical methods in
  natural language processing (EMNLP)}, pp.\  7534--7550, 2020.

\bibitem[Wang et~al.(2024)Wang, Downey, Ji, and Hope]{wang2024scimon}
Qingyun Wang, Doug Downey, Heng Ji, and Tom Hope.
\newblock Scimon: Scientific inspiration machines optimized for novelty.
\newblock In \emph{Proceedings of the 62nd annual meeting of the association
  for computational linguistics (volume 1: long papers)}, pp.\  279--299, 2024.

\bibitem[Wei et~al.(2026)Wei, Wang, Liao, Dong, Liu, Jia, Yu, and
  Zhu]{wei2026genprove}
Jingxuan Wei, Xingyue Wang, Yanghaoyu Liao, Jie Dong, Yuchen Liu, Caijun Jia,
  Bihui Yu, and Junnan Zhu.
\newblock Genprove: Learning to generate text with fine-grained provenance.
\newblock In \emph{Proceedings of the 64th Annual Meeting of the Association
  for Computational Linguistics (Volume 1: Long Papers)}, pp.\  5027--5048,
  2026.

\bibitem[Yuan et~al.(2026)Yuan, Kong, Li, Li, and Liu]{yuan2026confrag}
Yizhen Yuan, Rui Kong, Dongze Li, Yuanchun Li, and Yunxin Liu.
\newblock Benchmarking llm's capability in reasoning over conflicting web
  references.
\newblock In \emph{Proceedings of the 64th Annual Meeting of the Association
  for Computational Linguistics (Volume 1: Long Papers)}, pp.\  303--322, 2026.

\bibitem[{Z.ai}(2025)]{glm2025glm4model}
{Z.ai}.
\newblock {GLM-4-9B-0414: Official Model Card}.
\newblock Hugging Face model repository, 2025.
\newblock URL \url{https://huggingface.co/zai-org/GLM-4-9B-0414}.
\newblock Accessed 19 September 2026.

\bibitem[Zhang et~al.(2019)Zhang, Kishore, Wu, Weinberger, and
  Artzi]{zhang2020bertscore}
Tianyi Zhang, Varsha Kishore, Felix Wu, Kilian~Q Weinberger, and Yoav Artzi.
\newblock Bertscore: Evaluating text generation with bert.
\newblock \emph{arXiv preprint arXiv:1904.09675}, 2019.

\bibitem[Zhao et~al.(2019)Zhao, Peyrard, Liu, Gao, Meyer, and
  Eger]{zhao2019moverscore}
Wei Zhao, Maxime Peyrard, Fei Liu, Yang Gao, Christian~M Meyer, and Steffen
  Eger.
\newblock Moverscore: Text generation evaluating with contextualized embeddings
  and earth mover distance.
\newblock In \emph{Proceedings of the 2019 conference on empirical methods in
  natural language processing and the 9th international joint conference on
  natural language processing (EMNLP-IJCNLP)}, pp.\  563--578, 2019.

\bibitem[Zhao et~al.(2025)Zhao, Huang, Hu, Wang, Mao, Zhang, Jiang, Wu, Ai,
  Wang, et~al.]{zhao2025swift}
Yuze Zhao, Jintao Huang, Jinghan Hu, Xingjun Wang, Yunlin Mao, Daoze Zhang,
  Zeyinzi Jiang, Zhikai Wu, Baole Ai, Ang Wang, et~al.
\newblock Swift: a scalable lightweight infrastructure for fine-tuning.
\newblock In \emph{Proceedings of the AAAI Conference on Artificial
  Intelligence}, volume~39, pp.\  29733--29735, 2025.

\bibitem[Zheng et~al.(2023)Zheng, Chiang, Sheng, Zhuang, Wu, Zhuang, Lin, Li,
  Li, Xing, et~al.]{findlinkZheng2023Judging}
Lianmin Zheng, Wei-Lin Chiang, Ying Sheng, Siyuan Zhuang, Zhanghao Wu, Yonghao
  Zhuang, Zi~Lin, Zhuohan Li, Dacheng Li, Eric Xing, et~al.
\newblock Judging llm-as-a-judge with mt-bench and chatbot arena.
\newblock \emph{Advances in neural information processing systems},
  36:\penalty0 46595--46623, 2023.

\bibitem[Zheng et~al.(2026)Zheng, Huang, Yu, Yao, Ye, Chen, Luo, Ding, Bai,
  Cui, et~al.]{zheng2026sciverifier}
Shenghe Zheng, Chenyu Huang, Fangchen Yu, Junchi Yao, Jingqi Ye, Tao Chen, Yun
  Luo, Ning Ding, Lei Bai, Ganqu Cui, et~al.
\newblock Sci-verifier: Scientific verifier with thinking.
\newblock In \emph{International Conference on Learning Representations},
  volume 2026, pp.\  77463--77504, 2026.

\bibitem[Zhu et~al.(2025)Zhu, Xiao, Wang, Zhai, Zhou, and Zong]{zhu2025trove}
Junnan Zhu, Min Xiao, Yining Wang, Feifei Zhai, Yu~Zhou, and Chengqing Zong.
\newblock Trove: A challenge for fine-grained text provenance via source
  sentence tracing and relationship classification.
\newblock In \emph{Proceedings of the 63rd Annual Meeting of the Association
  for Computational Linguistics (Volume 1: Long Papers)}, pp.\  11755--11771,
  2025.

\end{thebibliography}
\bibliographystyle{iclr2027_conference}

\clearpage
\appendix
\clearpage
\providecommand{\flappendixsection}[1]{\FloatBarrier\section{#1}}
\providecommand{\flappendixcite}[1]{{\fontsize{7.4}{9}\selectfont\citep{#1}}}
\definecolor{FLPromptPaper}{HTML}{FAEFE9}
\definecolor{FLPromptBorder}{HTML}{D1C3BA}
\definecolor{FLPromptTitle}{HTML}{E7DAD1}
\definecolor{FLCasePaper}{HTML}{F6F8FA}
\definecolor{FLCaseBorder}{HTML}{CBD5DF}
\definecolor{FLCaseTitle}{HTML}{E5EBF1}
\newtcblisting{flpromptbox}[1]{
  enhanced,breakable,listing only,listing engine=listings,
  colback=FLPromptPaper,colframe=FLPromptBorder,
  colbacktitle=FLPromptTitle,coltitle=black,
  title={#1},fonttitle=\bfseries\small,
  boxrule=0.45pt,arc=1.2pt,left=7pt,right=7pt,top=5pt,bottom=5pt,
  before skip=7pt,after skip=9pt,
  listing options={basicstyle=\fontsize{9}{11}\selectfont\rmfamily,
    breaklines=true,columns=fullflexible,keepspaces=true,
    showstringspaces=false,literate={²}{{\textsuperscript{2}}}1,
    aboveskip=0pt,belowskip=0pt}
}
\newtcolorbox{flcasebox}[1]{
  enhanced,colback=FLCasePaper,colframe=FLCaseBorder,
  colbacktitle=FLCaseTitle,coltitle=black,title={#1},
  fonttitle=\bfseries\small,fontupper=\small,
  boxrule=0.45pt,arc=1.2pt,left=7pt,right=7pt,top=5pt,bottom=5pt,
  before skip=7pt,after skip=9pt
}
\setlength{\textfloatsep}{11pt plus 2pt minus 2pt}
\setlength{\floatsep}{9pt plus 2pt minus 2pt}
\setlength{\intextsep}{10pt plus 2pt minus 2pt}
\setlength{\abovecaptionskip}{5pt}
\setlength{\belowcaptionskip}{5pt}
\renewcommand{\topfraction}{0.95}
\renewcommand{\bottomfraction}{0.85}
\renewcommand{\textfraction}{0.05}
\renewcommand{\floatpagefraction}{0.75}
\setcounter{topnumber}{4}
\setcounter{bottomnumber}{3}
\setcounter{totalnumber}{6}

\flappendixsection{Additional Dataset Details}
\label{app:dataset_details}

\subsection{Task Definition and Evidence Representation}
\label{app:task_definition}
A GI-TraceBench instance is $(q,t,E,y^{\mathrm{ref}})$: a research question,
its task family, a supplied evidence packet, and a structured reference
answer. Systems receive $(q,t,E)$; the reference answer is withheld.
A \emph{Paper} identifies a source document, a \emph{Claim} records a
scientific statement with contextual qualifications, and an
\emph{Evidence Span} locates the associated source passage. Association
records and hypothesis cards refer to these objects when organizing
candidate connections and their assessment requirements.

\emph{Provenance QA} targets source localization, support judgments, and
study conditions. \emph{Knowledge Discovery QA} targets premise relations,
bounded conclusions, and missing evidence. \emph{Hypothesis Evaluation QA}
targets evidential grounding, predicted observations, verification
requirements, and refutation criteria. Required fields depend on the
question subtype; a candidate assessment is not an experimentally
established finding.

The evidence packets contain source text and may include public Claims,
contextual metadata, and source identifiers. The benchmark measures reasoning over supplied evidence, including the
public structured records available in each packet.

\subsection{Construction and Quality Control}
\label{app:construction_details}
Literature is organized into thematic clusters while retaining document
identifiers, section locations, and available passages. Full text,
abstract-only material, and bibliographic pointers are distinguished.
Parsing, evidence localization, and normalization yield source-linked
Claims; normalized concepts and topic constraints connect them into
candidate associations without removing their separate study contexts.
Association-based, typed-chain, and single-paper routes produce hypothesis
cards with predictions, verification requirements, refutation criteria,
and limitations.

QA records are assembled from these linked objects. Fused provenance
questions retain the source references of their component checks;
discovery and hypothesis questions reuse the associated Claims and spans.
Reference-closure checks ensure that linked identifiers resolve.
Supervision includes silver-label components derived through deterministic
transformations.

\paragraph{Human-in-the-loop release audit.}
Following construction, we manually audited a 20\% sample of the Train split
and 20\% of the Test split. Two academic reviewers with backgrounds in
computer technology checked every reference label in the sampled records
against the supplied evidence and task schema. All labels in the audited
sample were confirmed before the dataset was frozen. The released benchmark
therefore uses the post-audit reference labels.

\paragraph{Train/Test separation.}
\label{app:split_protocol}
The Test split comprises clusters G010, G018, G025, G031, G039, G041, G044, and G048; the remaining 42 clusters form Train. Overlap checks cover Paper, Claim,
Evidence Span, Association, and Hypothesis identifiers, including nested
references. Questions are normalized by Unicode NFKC, case folding,
whitespace collapse, and trimming. Test membership takes precedence when
resolving cross-partition conflicts. The retained splits have no overlap
in these checked identifiers or normalized questions. Semantic near-duplicates
and pretraining exposure remain possible. The complete 1,545-question Test split is used as the fixed evaluation set
throughout the reported automatic comparisons.

\subsection{Dataset Statistics}
\label{app:dataset_statistics}
Tables~\ref{tab:dataset_splits} and~\ref{tab:domain_distribution} give the
retained QA counts. Embedded checks and reused evidence objects do not
increase the number of QA instances. The construction record contains
10,733 source QA records and 451 exclusions before the retained release;
these are construction-stage counts, not a new Train/Test filtering run.
Table~\ref{tab:construction_inventory} reports upstream objects separately.

\begin{table}[!htbp]
    \centering
    \small
    \setlength{\tabcolsep}{4pt}
    \renewcommand{\arraystretch}{1.10}
    \caption{\textbf{Train/Test composition.} Counts refer to retained QA instances and thematic clusters.}
    \label{tab:dataset_splits}
    \begin{tabular*}{\linewidth}{@{\extracolsep{\fill}}lrrrrr@{}}
    \toprule
    Split & Clusters & Provenance & Discovery & Hypothesis & Total \\
    \midrule
    Train & 42 & 5,009 & 1,624 & 2,104 & 8,737 \\
    Test & 8 & 898 & 286 & 361 & 1,545 \\
    \midrule
    \textbf{Total} & \textbf{50} & \textbf{5,907} & \textbf{1,910}
        & \textbf{2,465} & \textbf{10,282} \\
    \bottomrule
    \end{tabular*}
\end{table}
\begin{table}[!htbp]
    \centering
    \small
    \renewcommand{\arraystretch}{1.10}
    \caption{\textbf{Domain distribution.} Cross-GI denotes shared gastrointestinal oncology topics, not a separate cancer site.}
    \label{tab:domain_distribution}
    \begin{tabular*}{\linewidth}{@{\extracolsep{\fill}}lrrr@{}}
    \toprule
    Domain & Train & Test & Total \\
    \midrule
    Biliary & 300 & 31 & 331 \\
    Colorectal & 2,157 & 560 & 2,717 \\
    Cross-GI & 2,380 & 596 & 2,976 \\
    Esophageal & 973 & 221 & 1,194 \\
    Gastric & 1,380 & 83 & 1,463 \\
    Pancreatic & 1,547 & 54 & 1,601 \\
    \midrule
    \textbf{Total} & \textbf{8,737} & \textbf{1,545}
        & \textbf{10,282} \\
    \bottomrule
    \end{tabular*}
\end{table}
\begin{table}[!htbp]
    \centering
    \small
    \setlength{\tabcolsep}{4pt}
    \renewcommand{\arraystretch}{1.10}
    \caption{\textbf{Upstream construction inventory.} These counts precede formal QA filtering and are not split-level entity counts.}
    \label{tab:construction_inventory}
    \begin{tabularx}{\linewidth}{@{}l r >{\raggedright\arraybackslash}X@{}}
    \toprule
    Object or construction route & Count & Counting unit / scope \\
    \midrule
    Papers & 1,614 & Source-document records. \\
    Claims & 5,475 & Scientific-statement records. \\
    Evidence Spans & 4,968 & Source-passage records. \\
    Associations & 733 & Candidate-association records. \\
    \midrule
    Hypothesis cards: association-based & 733 & Cards derived from association records. \\
    Hypothesis cards: typed-chain & 838 & Cards derived from typed discovery chains. \\
    Hypothesis cards: single-paper & 477 & Cards derived from single-paper records. \\
    \textbf{Hypothesis cards: total} & \textbf{2,048} & Sum of the three card-construction routes. \\
    \bottomrule
    \end{tabularx}
    \par\vspace{3pt}
    \begin{minipage}{\linewidth}\small
    A card can support multiple questions. The 2,048 construction-stage
    cards must not be equated with the 2,465 retained Hypothesis Evaluation
    QA instances. Different object types are not additive counting units.
    \end{minipage}
\end{table}
\FloatBarrier

\subsection{Benchmark Comparison}
\label{app:benchmark_comparison}
Table~\ref{tab:benchmark_comparison} compares capabilities covered by
specified benchmark tasks and their reference or evaluation targets.
Auxiliary construction outputs alone do not establish capability coverage. A check requires coverage
in at least one specified task or subset; a cross means that the capability
is not explicitly covered under the definitions below. The comparison
concerns benchmark design, not model performance or annotation quality.

\begin{table}[!htbp]
    \centering
    \fontsize{8.2}{10}\selectfont
    \setlength{\tabcolsep}{2pt}
    \renewcommand{\arraystretch}{1.10}
    \caption{\textbf{Capability coverage.} \gicmark: explicitly covered; \gixmark: not explicitly covered. Coverage is benchmark-level, not required of every instance.}
    \label{tab:benchmark_comparison}
    \begin{tabular*}{\linewidth}{@{\extracolsep{\fill}}l*{10}{c}@{}}
    \toprule
    & \multicolumn{5}{c}{Evidence grounding}
    & \multicolumn{3}{c}{Scientific reasoning}
    & \multicolumn{2}{c}{Further testing} \\
    \cmidrule(lr){2-6}\cmidrule(lr){7-9}\cmidrule(lr){10-11}
    Benchmark & SA & EL & MD & CV & UI & SC & SR & HA & VP & RC \\
    \midrule
    PubMedQA~\flappendixcite{jin2019pubmedqa} & \gixmark & \gixmark & \gixmark & \gixmark & \gicmark & \gixmark & \gixmark & \gixmark & \gixmark & \gixmark \\
    BioASQ-QA~\flappendixcite{krithara2023bioasqqa} & \gicmark & \gicmark & \gicmark & \gixmark & \gixmark & \gixmark & \gixmark & \gixmark & \gixmark & \gixmark \\
    QASPER~\flappendixcite{dasigi2021qasper} & \gicmark & \gicmark & \gixmark & \gixmark & \gicmark & \gixmark & \gixmark & \gixmark & \gixmark & \gixmark \\
    SciFact~\flappendixcite{wadden2020scifact} & \gicmark & \gicmark & \gicmark & \gicmark & \gicmark & \gixmark & \gixmark & \gixmark & \gixmark & \gixmark \\
    ScholarQABench~\flappendixcite{asai2026synthesizing} & \gicmark & \gicmark & \gicmark & \gicmark & \gixmark & \gixmark & \gixmark & \gixmark & \gixmark & \gixmark \\
    TROVE~\flappendixcite{zhu2025trove} & \gicmark & \gicmark & \gicmark & \gixmark & \gixmark & \gixmark & \gixmark & \gixmark & \gixmark & \gixmark \\
    ReFInE (GenProve)~\flappendixcite{wei2026genprove} & \gicmark & \gicmark & \gicmark & \gixmark & \gixmark & \gixmark & \gixmark & \gixmark & \gixmark & \gixmark \\
    ResearchPulse-Bench~\flappendixcite{chen2025researchpulse} & \gicmark & \gixmark & \gicmark & \gixmark & \gixmark & \gicmark & \gicmark & \gixmark & \gixmark & \gixmark \\
    \midrule
  \multicolumn{11}{@{}l}{\emph{Discovery and hypothesis comparators}} \\
  ResearchBench~\flappendixcite{liu2026researchbench}
& \gixmark & \gixmark & \gicmark & \gixmark & \gixmark
& \gixmark & \gicmark & \gicmark & \gixmark & \gixmark \\
    DiscoveryBench~\flappendixcite{majumder2024discoverybench} & \gixmark & \gixmark & \gixmark & \gixmark & \gixmark & \gicmark & \gicmark & \gicmark & \gixmark & \gixmark \\
    \midrule
    \textbf{GI-TraceBench (ours)} & \gicmark & \gicmark & \gicmark & \gicmark & \gicmark & \gicmark & \gicmark & \gicmark & \gicmark & \gicmark \\
    \bottomrule
    \end{tabular*}

\end{table}

\paragraph{Capability definitions.}
\textbf{SA} links an answer or claim to sources used as evidential
support; inspiration retrieval alone does not qualify;
\textbf{EL} is subdocument evidence localization;
\textbf{MD} links a target to multiple source documents;
\textbf{CV} gives evidence-based verdicts on supplied claims;
\textbf{UI} gives uncertainty or insufficient-evidence status.
\textbf{SC} explicitly annotates or evaluates preservation of
study-context boundaries; \textbf{SR} represents scientific relations among findings,
methods, or variables. \textbf{HA} evaluates candidate hypotheses;
\textbf{VP} specifies further observations or tests needed to establish a
candidate; \textbf{RC} specifies outcomes that would undermine it.

\paragraph{Comparison scope.}
PubMedQA's \emph{maybe} label supports UI, not claim verification by itself
\citep{jin2019pubmedqa}. ScholarQABench is assessed across its specified
subsets; it does not inherit excluded uncertainty labels from its source
datasets \citep{asai2026synthesizing}. The provenance-relation tags in TROVE and ReFInE do not alone constitute the scientific relations denoted by SR
\citep{zhu2025trove,wei2026genprove}. ResearchPulse's context and relation
coverage refers to aligned experimental targets and method relations
\citep{chen2025researchpulse}. ResearchBench evaluates inspiration retrieval, hypothesis composition,
and pairwise hypothesis ranking. Its inspiration links are not scored as
evidential support attribution, and study-context boundary preservation is
not an explicit evaluation target. Its extracted experimental-design fields
are auxiliary construction outputs and are outside the three evaluated
tasks \citep{liu2026researchbench}. DiscoveryBench uses empirical datasets,
rather than literature evidence packets, for hypothesis search and
verification \citep{majumder2024discoverybench}. GI-TraceBench covers the ten
selected capabilities across its three task families without claiming
exhaustive sentence-level provenance supervision.

\flappendixsection{Method and Implementation Details}
\label{app:method_details}

\subsection{Canonical Evidence and Premise Contract}
\label{app:premise_contract}

The canonical evidence view preserves the source objects supplied to the
system. Each evidence record contains a paper identifier, a claim identifier,
an evidence span, its local position, and an availability indicator:
\begin{equation}
V(E)=
\left\{
r_j=(p_j,c_j,s_j,\ell_j,a_j)
\right\}_{j=1}^{|E|},
\label{eq:canonical_evidence_view}
\end{equation}
where $p_j$ identifies a paper, $c_j$ identifies a claim, $s_j$ is an
evidence span, $\ell_j$ records its section or local position, and $a_j$
records evidence availability.

Premise Grounding maps this evidence view to an ordered inventory
$\mathcal P$ and a limitation set $\Lambda$. The premise fields are
summarized in Table~\ref{tab:premise_contract}.

\begin{table}[!htbp]
\centering
\renewcommand{\arraystretch}{1.10}
\small
\caption{\textbf{Premise representation.}
Support labels are model predictions, not independently verified truth labels.}
\label{tab:premise_contract}
\begin{tabularx}{\linewidth}{@{}l l >{\raggedright\arraybackslash}X@{}}
\toprule
Field & Notation & Meaning \\
\midrule
\texttt{id} & $i$ & Local premise identifier, such as \texttt{P0}. \\
\texttt{statement} & $s_i$ & Atomic statement to be assessed or reused. \\
\texttt{context} & $c_i$ & Study setting and applicable qualifications. \\
\texttt{polarity} & $\pi_i$ & Positive or negative orientation. \\
\texttt{origin} & $o_i$ & Observation, author claim, or candidate origin. \\
\texttt{sources} & $I_i$ & References to canonical evidence pointers. \\
\texttt{support} & $\sigma_i$ & Predicted support judgment. \\
\bottomrule
\end{tabularx}
\end{table}

The origin field distinguishes source observations, author claims, derived
candidates, and question-introduced candidates. Support takes one of the
values \texttt{supported}, \texttt{refuted}, \texttt{partial}, or
\texttt{not\_established}. Polarity is separate from support: a negative
statement may be supported, and a positive statement may be refuted.

Each source pointer $I_i$ refers to a record in $V(E)$. The premise inventory
is not required to be exhaustive. An empty or incomplete inventory does not
imply that the original evidence is absent; later reasoning may still use a
direct source binding.

\subsection{Support-Plan Interface}
\label{app:plan_language}

The support plan is a typed, acyclic representation of how supplied evidence
is used for the requested assessment. The plan contains a target node,
statement nodes, and relation nodes, and is represented in the main text as
$T=(r,\mathcal N,\mathcal R)$.

The support-plan interface provides three selectors and several control
fields:

\begin{table}[!htbp]
\centering
\renewcommand{\arraystretch}{1.10}
\small
\caption{\textbf{Support-plan interface.}
The operators organize a represented assessment and do not constitute a
general scientific proof system.}
\label{tab:plan_interface}

\begin{tabularx}{\linewidth}{
@{}p{0.27\linewidth}
>{\raggedright\arraybackslash}X@{}
}
\toprule
Interface & Role and interpretation \\
\midrule

\texttt{from\_a} &
Selects a generated premise by local identifier and retains its source
bindings. \\

\texttt{from\_source} &
Binds a proposition directly to an original evidence pointer. \\

\texttt{all\_of} &
Combines existing support expressions without introducing a new relation. \\

Context metadata &
Preserves study setting, polarity, and scope. \\

Target requirements &
Records requested propositions and concrete verification conditions. \\

Validity judgment &
Records whether a candidate relation is admissible as a represented
assessment. \\

\shortstack[l]{\texttt{representation\_}\\
\texttt{unsupported}} &
Records a limitation of the restricted representation separately from
insufficient scientific evidence. \\

\bottomrule
\end{tabularx}
\end{table}

The schema $\mathcal G(x,A)$ restricts available identifiers, expression
shapes, depth, and leaf count. Every relation has nonempty inputs, an output,
a validity judgment, and specific conditions. Existing roots cannot be
overwritten. A correction is allowed only when it is explicitly grounded in
an available evidence pointer.

The validity judgment is a fallible model judgment rather than a program
certificate. A plan may therefore preserve supported observations together
with a conditional or unestablished relation.

\subsection{Deterministic Compilation}
\label{app:compiler_details}

The compiler is non-learned. It validates the plan, resolves identifiers,
restores selected evidence and citation bindings, records premise use, and
computes the fields passed to the final-answer schema.

At the representation level, a plan is admissible when
\begin{equation}
\nu_T=
\mathbf{1}\!\left[
\operatorname{Acyclic}(T)
\land
\operatorname{Ids}_A(T)\subseteq\operatorname{Ids}(A)
\land
\operatorname{Ids}_E(T)\subseteq\operatorname{Ptr}(V(E))
\land
\operatorname{Shape}(T)\in\mathcal G(x,A)
\right].
\label{eq:appendix_compile_validity}
\end{equation}

For a plan $T$, the premise records selected through
\texttt{from\_a} are
\begin{equation}
U_A(T)=
\{i:\texttt{from\_a}(i)\text{ occurs in }T\},
\qquad
M_A(T)=
\operatorname{Ids}(\mathcal P)\setminus U_A(T).
\label{eq:premise_consumption}
\end{equation}

A source selected through \texttt{from\_source} does not need to occur in
$U_A(T)$. The compiler therefore distinguishes premise consumption from
direct access to original evidence.

The compiler record contains the structural validity flag, restored evidence,
consumed premise identifiers, remaining conditions, and represented support
status:
\begin{equation}
C=
\Gamma_{\mathcal G}(V(E),A,T)
=
\bigl(\nu_T,\widehat E_T,U_A(T),K_T,\eta_T\bigr).
\label{eq:appendix_compile_output}
\end{equation}

Here, $\widehat E_T$ is the restored evidence record, $K_T$ contains
remaining conditions, and $\eta_T$ records the represented status. The status
is an assessment of the supplied representation rather than a certificate of
scientific truth.

Conditions in $K_T$ are concrete and testable. Alternative support paths
remain alternatives rather than being collapsed into one conjunction.
Repeated source identifiers do not independently discharge a condition.
These checks maintain structural and source-binding consistency, while
scientific entailment remains dependent on the model's interpretation of the
evidence.

\subsection{Supervision and Training Configuration}
\label{app:method_training_details}
\label{app:method_supervision}

The three supervision sets for sequential adaptation are:

\begin{itemize}
\item $\mathcal D_A$ contains the question, task family, canonical evidence
view, and source-bound premise target;
\item $\mathcal D_{\mathrm{plan}}$ contains the question, evidence view,
premise inventory, and support-plan target;
\item $\mathcal D_{\mathrm{answer}}$ contains the question, evidence view,
premise inventory, support plan, compiler record, and structured answer target.
\end{itemize}

The expanded objectives in the main text are implemented with
assistant-response likelihood:
\begin{equation}
\mathcal L(\theta;\mathcal D)
=
-\sum_{(u,v)\in\mathcal D}
\sum_{\ell\in\mathcal M(v)}
\log p_\theta(v_\ell\mid u,v_{<\ell}),
\label{eq:findlink_training_loss}
\end{equation}
where $u$ is the stage input and $\mathcal M(v)$ selects assistant-response
tokens. Evidence, premises, and compiler feedback are conditioning contexts
rather than additional loss targets.

Premise adaptation uses $\mathcal D_A$. Reasoning adaptation uses both
$\mathcal D_{\mathrm{plan}}$ and $\mathcal D_{\mathrm{answer}}$. Empty-premise
conditions are included during reasoning supervision so that the model learns
that an incomplete premise inventory does not imply missing evidence.

\paragraph{Optimization.}
\label{app:method_hyperparameters}

Qwen3.5-9B \citep{qwen2026qwen35model} is adapted with LoRA
\citep{hu2021lora} using MS-SWIFT \citep{zhao2025swift}. The route is
$S\rightarrow SA\rightarrow SAB$, where $S$ is ordinary SFT, $SA$ is the
premise-adapted checkpoint, and $SAB$ is the shared reasoning checkpoint.
The backbone parameters remain frozen during LoRA adaptation, and each stage
starts a new optimizer.

\begin{table}[!htbp]
\centering
\renewcommand{\arraystretch}{1.10}
\small
\caption{\textbf{Configuration of the two adaptation stages.}}
\label{tab:method_hyperparameters}
\begin{tabularx}{\linewidth}{@{}>{\raggedright\arraybackslash}X l l@{}}
\toprule
Setting & Premise adaptation & Reasoning adaptation \\
\midrule
Initialization & $S$ & $SA$ \\
Targets & Premises & Plans and answers \\
Adaptation & LoRA & LoRA \\
LoRA rank / scaling alpha & 16 / 32 & 16 / 32 \\
Epochs & 1 & 1 \\
Learning rate & $10^{-4}$ & $10^{-4}$ \\
Warmup ratio & 0.05 & 0.05 \\
Precision & bf16 & bf16 \\
Random seed & 42 & 42 \\
Maximum sequence length & 32,768 & 32,768 \\
Micro-batch per device & 1 & 1 \\
Gradient accumulation & 4 & 2 \\
Effective global batch & 12 & 6 \\
Optimizer state & New for stage & New for stage \\
\bottomrule
\end{tabularx}
\end{table}

\subsection{Inference and Ablation Interfaces}
\label{app:method_inference}
\label{app:method_decoding}

We use greedy decoding with temperature 0, top-$p$ 1, and thinking disabled.
XGrammar 0.2.5.post1 enforces the output schemas
\citep{dong2024xgrammar}. Requested new-token budgets are 2,048 for
premises, 2,048 for planning, and 4,096 for answering.

For prompt length $L_s<32{,}768$, the effective token budget is
\begin{equation}
b_s^{\mathrm{eff}}
=
\min\{b_s,32{,}768-L_s\}.
\label{eq:findlink_budget}
\end{equation}

Inputs are not silently truncated. Invalid generation or invalid
representation is recorded as an execution outcome, rather than as evidence
that a scientific claim is false. The compiler and final audit make no model
calls.

\begin{algorithm}[!htbp]
\caption{FindLink inference over supplied evidence}
\label{alg:findlink_inference}
\small
\begin{algorithmic}[1]
\Require Question $q$, task family $t$, evidence $E$,
checkpoints $\theta_A,\theta_B$
\State $x\gets(q,t,V(E))$
\State $A\gets\operatorname{Ground}_{\theta_A}(x)$
\State If $A$ is empty or incomplete, retain $V(E)$ and allow
       \texttt{from\_source} leaves
\State $T\gets\operatorname{Plan}_{\theta_B}(x,A)$
\State $C\gets\operatorname{Compile}(x,A,T)$
\State Return an execution outcome if compilation does not permit answering
\State $\widetilde y\gets\operatorname{Answer}_{\theta_B}(x,A,T,C)$
\State $\hat y\gets\operatorname{Audit}_{\mathcal S_t}(\widetilde y)$
\State \Return $\hat y$ and the recorded outcome
\end{algorithmic}
\end{algorithm}

\paragraph{Ablation interfaces.}
\label{app:ablation_details}
\label{app:ablation_judge}

For w/o A, \texttt{from\_a} leaves are disabled and every plan leaf binds
directly to original evidence. Compilation remains active. For w/o B, the
premise-adapted checkpoint is retained, but support planning and
compiler-derived feedback are removed. FindLink uses the complete route
$S\rightarrow SA\rightarrow SAB$.

\begin{table}[!htbp]
\centering
\small
\caption{\textbf{Core comparison design.}}
\label{tab:ablation_design}
\begin{tabularx}{\linewidth}{@{}>{\raggedright\arraybackslash}p{0.18\linewidth}
p{0.20\linewidth}>{\raggedright\arraybackslash}X@{}}
\toprule
Configuration & Training route & Inference interface \\
\midrule
Default SFT & Reuse $S$ &
Question and evidence directly produce the answer. \\
w/o A & $S\rightarrow SB$ &
Plan and answer directly from original evidence; no premise inventory or
\texttt{from\_a} leaves. \\
w/o B & $SA\rightarrow D$ &
Frozen premise output plus question and evidence; no support plan or
compiler-derived fields. \\
FindLink & $S\rightarrow SA\rightarrow SAB$ &
Premises, support plan, deterministic compilation, and
evidence-conditioned answer. \\
\bottomrule
\end{tabularx}
\end{table}

The complete shared instruction, Premise Grounding prompt, Structured
Reasoning prompt, compiler-conditioned answer prompt, and LLM-as-Judge
prompt are provided in Appendix~\ref{app:prompt_findlink_stages}. These
prompts instantiate the interfaces defined above but do not alter the
deterministic compiler or evaluation metrics.

\flappendixsection{Evaluation Protocol}
\label{app:experimental_protocol}

\subsection{Evaluation Scope and Baseline Adaptation}
\label{app:experiment_inputs}
For each evaluated question, systems receive the same cached evidence;
reference answers are withheld. All 19 configurations use the complete 1,545-question Test split for the
reported task-specific and text-similarity metrics. All 19 systems are evaluated by the LLM judge on the same 120 questions,
with 40 questions from each task family. Human evaluation uses a shared
24-question subset, yielding 456 system--question pairs. For Human--LLM
correlation, both judge and human scores are restricted to the same
24-question subset and include the 11 open- and closed-source comparison
models plus FindLink, yielding 288 pairs summarized as 12 system means.

\paragraph{Model and method references.}
\label{app:model_references}
\label{app:baseline_adaptation}
Model names and original release references are retained in the result
tables. Attribute First \citep{slobodkin2024attribute}, OpenScholar
\citep{asai2026synthesizing}, PaperQA2 \citep{skarlinski2024paperqa2}, and
VeriCite \citep{qian2025vericite} are adapted to the common supplied-evidence
input/output setting. Their rows are local adaptations, not reproductions
of their original end-to-end retrieval systems. Backbones, in-domain
supervision, and inference budgets are not matched across the comparison.

\paragraph{Output processing.}
\label{app:experiment_normalization}
Uniquely recoverable syntax and wrapper compatibility are normalized
without another model call or reference-derived answer content. Failed
predictions remain in their applicable denominators. Output-format compliance is assessed after processing; adapters
can wrap natural-language answers and uniquely mappable citations.
Structured extraction is not guaranteed to treat every semantically
equivalent answer identically. The reported results are point estimates from the recorded runs.

\subsection{Task Metrics and Diagnostic Components}
\label{app:experimental_metrics}
\label{app:task_metric_definitions}
Let $\overline m_k$ denote a task-family component after scoring within
applicable clusters and then macro-averaging across clusters. Claim-verdict
and sufficiency classification F1 use pooled labels within each cluster,
not a simple average of per-question classification F1. The frozen task
scores are
\begin{align}
\mathrm{PGS}&=\tfrac13(\overline m_{\mathrm{span\_token\_f1}}
+\overline m_{\mathrm{claim\_verdict\_f1}}+\overline m_{\mathrm{scope\_slot\_f1}}),
\label{eq:pgs_definition}\\
\mathrm{DRS}&=\tfrac15(\overline m_{\mathrm{premise\_f1}}
+\overline m_{\mathrm{relation\_f1}}+\overline m_{\mathrm{boundary\_f1}}
+\overline m_{\mathrm{sufficiency\_f1}}+\overline m_{\mathrm{gap\_coverage}}),
\label{eq:drs_definition}\\
\mathrm{HSS}&=0.30\overline m_{\mathrm{ground\_f1}}
+0.15\overline m_{\mathrm{premise\_closure}}
+0.15\overline m_{\mathrm{verify\_plan}}\nonumber\\
&\quad+0.15\overline m_{\mathrm{refute\_comp}}
+0.15\overline m_{\mathrm{boundary\_comp}}
+0.10\overline m_{\mathrm{sufficiency\_f1}}.
\label{eq:hss_definition}
\end{align}
BFM remains $(\mathrm{PGS}+\mathrm{DRS}+\mathrm{HSS})/3$. Text and judge
scores are reported separately; no weights change for ablations.
HSS evaluates reference assessment content, not experimental success or
scientific novelty.

\paragraph{Four diagnostic F1 scores.}
Premise F1 matches \texttt{premise\_claim\_ids}, with a fallback to
\texttt{claim\_ids} when the reference premise set is empty. Relation F1
matches association identifiers and relation types extracted from
\texttt{relation\_family}, \texttt{association\_type}, and
\texttt{predicate}. Boundary F1 compares normalized context
\texttt{key=value} slots. These set scores are averaged within clusters,
then across clusters. Sufficiency F1 uses rule-derived
\texttt{answerable}, \texttt{conservative}, and \texttt{abstain} labels;
class F1 is averaged for labels present in each cluster's references,
followed by cluster averaging. The set scorer assigns 1 when both extracted
sets are empty. Thus, these metrics measure extractable reference-bound
content, not unrestricted semantic equivalence. Gap coverage remains part
of DRS but is not plotted as an F1 score.

\subsection{Text Metrics}
\label{app:text_metric_details}
ROUGE-L \citep{lin2004rouge}, METEOR \citep{banerjee2005meteor},
BERTScore-F1 \citep{zhang2020bertscore}, and MoverScore
\citep{zhao2019moverscore} use a common rendered answer representation rather
than raw JSON. BERTScore uses \texttt{roberta-base}
\citep{liu2019roberta}, layer 10, without IDF or baseline rescaling, and a
512-token input limit. Each metric is computed within a task family and then
macro-averaged across the three families:
\begin{equation}
M_{\mathrm{family\text{-}macro}}=\tfrac13
\sum_{f\in\{\mathrm{P,D,H}\}}M_f.
\label{eq:metric_family_macro}
\end{equation}
Each $M_f$ is computed using the metric-specific aggregation within family
$f$ before the three family scores are averaged.

\subsection{LLM-as-Judge and Human Evaluation}
\label{app:llm_judge}
The fixed anonymous judge uses only visible evidence. The reference answer
specifies requested content and omissions, not additional factual evidence.
For each response, it returns one holistic score in
$\{0,0.5,\ldots,5\}$ and a short rationale. It considers the applicable
criteria in Table~\ref{tab:judge_dimensions} equally, but does not output
per-dimension scores. The system score for the main LLM-as-Judge evaluation is the mean over
the shared 120 questions. The complete instruction appears in
Appendix~\ref{app:prompt_judge}.
\begin{table}[!htbp]
\centering\small
\caption{\textbf{Shared evaluation criteria.} The LLM judge assigns one holistic score; human reviewers score each applicable dimension. Citation criteria apply only when citations are required.}
\label{tab:judge_dimensions}
\begin{tabularx}{\linewidth}{@{}p{0.26\linewidth}X@{}}
\toprule
Criterion & Assessment \\
\midrule
Answer quality & Correctness, relevance, completeness, and coverage of the requested content. \\
Evidence entailment & Whether substantive claims follow from the supplied evidence without contradiction or unsupported extrapolation. \\
Citation correctness & Whether valid source references support the claims to which they are attached. \\
Citation completeness & Whether important evidence-dependent claims have the required source support. \\
Boundary control & Correct polarity, study scope, uncertainty, causal strength, and distinction between candidates, hypotheses, and facts. \\
\bottomrule
\end{tabularx}
\end{table}

Scores range from 5 (fully correct and complete) to 4 (minor limitations),
3 (usable with substantive omissions or partial errors), 2 (major errors),
1 (mostly unusable), and 0 (absent or invalid). Unsupported core claims,
severe boundary errors, or invalid core citations normally cap the score
at 2. Missing major required content prevents 5; justified evidence-bounded
abstention can be correct.

\paragraph{Human scoring.}
\label{app:human_llm_correlation}
Three reviewers independently score anonymized responses on each
applicable criterion in 0.5-point steps. Citation-requiring questions use
all five criteria; others use answer quality, evidence entailment, and
boundary control. Invalid answers receive zero. For system $m$, question
$q$, and applicable dimensions $D_q$,
\begin{equation}
H_{m,q}=\frac1{|D_q|}\sum_{d\in D_q}
\frac{h^{(1)}_{m,q,d}+h^{(2)}_{m,q,d}+h^{(3)}_{m,q,d}}3,\qquad
\overline H_m=\frac1{24}\sum_{q=1}^{24}H_{m,q}.
\label{eq:human_overall}
\end{equation}
A binary core-error label is recorded separately, not added to the score.
For each system--question pair, the final human score is obtained by
averaging the scores from the three independent reviewers. Agreement results are reported in
Appendix~\ref{app:human_agreement_results}.

\flappendixsection{Additional Analyses}
\label{app:extended_results}

\subsection{In-Domain Discovery Components}
\label{app:discovery_f1_all}

Figure~\ref{fig:diag_discovery_f1} reports the four Knowledge Discovery QA
components for the 16 comparison systems shown in the main text.
Table~\ref{tab:discovery_f1_all} complements that figure with Default SFT and
the two retrained ablations, together with FindLink as the common reference.
All four configurations are evaluated on the complete 286-question discovery
subset spanning eight thematic clusters. These values are frozen task-family
summaries, not averages over the 24-question qualitative subset.

\begin{table}[H]
\centering
\fontsize{8.4}{9.5}\selectfont
\setlength{\tabcolsep}{4.5pt}
\renewcommand{\arraystretch}{1.05}
\caption{\textbf{Knowledge Discovery QA components for the in-domain
configurations.} Values use the 0--100 scale. Bold marks the column maximum
among these four configurations. Results for the other 15 comparison systems
are displayed in Figure~\ref{fig:diag_discovery_f1}.}
\label{tab:discovery_f1_all}
\begin{tabularx}{\linewidth}{@{}>{\raggedright\arraybackslash}Xrrrr@{}}
\toprule
System & Premise & Relation & Boundary & Sufficiency \\
\midrule
Default SFT~\flappendixcite{qwen2026qwen35model}
& 59.93 & 58.69 & 61.63 & 62.06 \\
w/o A & 86.01 & 87.54 & 89.25 & 65.79 \\
w/o B & \textbf{88.10} & 76.44 & 89.74 & 60.71 \\
\rowcolor{FindLinkOurs}
\textbf{FindLink} & 87.69 & \textbf{88.98} & \textbf{92.36} & \textbf{78.40} \\
\bottomrule
\end{tabularx}
\end{table}

The component scores show that the full framework has the strongest relation,
boundary, and sufficiency scores among the four in-domain configurations,
whereas w/o B has the highest premise F1. Component definitions, empty-set
handling, and cluster-balanced aggregation are specified in
Appendix~\ref{app:task_metric_definitions}.
\FloatBarrier

\subsection{Human Evaluation Results}
\label{app:human_agreement_results}

For the Human--LLM agreement analysis, Table~\ref{tab:human_judge_results}
reports the means for 12 systems: the 11 open- and closed-source comparison
models and FindLink. All systems use the same pre-specified 24-question
subset for both judge and human scores. Adapted methods, Default SFT, and
the two ablations are excluded from this correlation analysis. Pearson
$r=0.958$ (Fisher 95\% CI $[0.854,0.989]$), Spearman
$\rho=0.928$, and $R^2=0.918$ describe a strong system-level association.
The mean absolute difference is 0.35 points, and 75.0\% of system means
differ by at most 0.5. These statistics use unrounded means.

Across the full 456-pair human pool, the three reviewers' overall-score
agreement is summarized by Spearman correlation 0.957, ordinal
Krippendorff's $\alpha$ 0.836, and ICC(2,1) 0.997 (two-way random effects,
absolute agreement, single measurement). Core-error agreement and Cohen's
$\kappa$ are both 1.000. These are reviewer-agreement statistics for the full
pool, distinct from the 12-model Human--LLM correlation. High correlation does
not imply item-level interchangeability; for example, Hunyuan-7B receives 3.48
from reviewers and 2.25 from the judge.

\begin{table}[H]
\centering
\fontsize{8.2}{9.3}\selectfont
\setlength{\tabcolsep}{4pt}
\renewcommand{\arraystretch}{1.00}
\caption{\textbf{Aligned human and LLM-judge means.} Each row reports a
system mean on the common 24-question subset and the 0--5 scale. The table
includes the 11 comparison models and FindLink; the judge values are
subset-specific for Human--LLM agreement, whereas the main table reports the
separate 120-question judge evaluation.}
\label{tab:human_judge_results}
\begin{tabular}{@{}>{\raggedright\arraybackslash}p{0.51\linewidth}*{2}{>{\raggedleft\arraybackslash}p{\dimexpr0.49\linewidth/2-2\tabcolsep\relax}}@{}}
\toprule
System & LLM Judge & Human \\
\midrule
\multicolumn{3}{@{}l}{\emph{Proposed method}} \\
\textbf{FindLink (Ours)} & \textbf{4.19} & \textbf{4.40} \\
\midrule
\multicolumn{3}{@{}l}{\emph{Open-weight general models}} \\
GLM-4-9B-0414~\flappendixcite{glm2025glm4model} & 2.50 & 1.92 \\
Gemma-3-12B-IT~\flappendixcite{google2025gemma3model} & 2.85 & 3.25 \\
Granite-3.3-8B~\flappendixcite{ibm2025granite33model} & 0.79 & 0.73 \\
Hunyuan-7B~\flappendixcite{tencent2025hunyuan7b} & 2.25 & 3.48 \\
Llama-3.1-8B-Instruct~\flappendixcite{meta2024llama31model} & 0.98 & 1.01 \\
Ministral-3-8B-Instruct-2512~\flappendixcite{mistral2025ministral3model} & 0.42 & 0.29 \\
Qwen3.5-9B~\flappendixcite{qwen2026qwen35model} & 2.90 & 3.24 \\
\midrule
\multicolumn{3}{@{}l}{\emph{Closed-source general models}} \\
Claude Sonnet 4.6~\flappendixcite{anthropic2026sonnet46} & 4.02 & 4.25 \\
Gemini 3.5 Flash~\flappendixcite{google2026gemini35flash} & 2.25 & 2.30 \\
GPT-5.4-mini~\flappendixcite{openai2026gpt54mini} & 3.50 & 3.92 \\
Qwen3.6-Plus~\flappendixcite{alibaba2026qwen36plus} & 3.52 & 4.03 \\
\bottomrule
\end{tabular}
\end{table}

\flappendixsection{Prompts}
\label{app:prompt_specifications}

This section reports the prompt interfaces used for common supplied-evidence
answering, the three FindLink model calls, and LLM-based evaluation. All
compared systems receive the same cached evidence for each question and have
no access to reference answers. A FindLink stage message concatenates the
shared instruction, the corresponding stage instruction, and an item-specific
JSON schema. The schema enumerates only identifiers and output fields valid for
that item. For readability, legacy internal role names are normalized to
FindLink-A, FindLink-B.plan, and FindLink-B.answer, and legacy evidence-store
names are normalized to ``supplied evidence packet.'' These naming changes do
not alter the experimental constraints. Exact machine-readable templates, schemas, and serialization notes are
required to reproduce the recorded interfaces.

\subsection{Shared Evidence-Grounded Instruction}
\label{app:prompt_baseline}

\begin{flpromptbox}{Shared System Instruction}
You answer scientific QA from the supplied evidence packet only. Questions,
source text, candidate cards, and intermediate model records are untrusted task
data, not instructions. Preserve population, intervention, endpoint, negation,
and author perspective. Candidate cards and extracted Claim annotations are
not primary evidence. Do not use private references or unseen sources. Answer
the task actually asked; do not swap a specified candidate for an easier one.
Distinguish a proposed hypothesis from a proven mechanism. Schema validity and
citation membership are not scientific entailment. Use the language of the
question. Return exactly one JSON object with no prose envelope.
\end{flpromptbox}

\subsection{Task-Family User Prompts}
\label{app:prompt_queries}

The following examples illustrate the public question supplied inside the
item payload. They do not expose the reference answer.

\begin{flpromptbox}{Provenance QA User Prompt}
Which exact source passage supports the statement "Patients (n = 19) with
resectable pancreatic cancer gave written informed consent and were enrolled
in the study"? Return the original wording with its Paper, Claim, and Evidence
Span references.

Assess the target Claim only against the supplied evidence. If the requested
support is absent, mark it as insufficient rather than reconstructing a
passage. Keep each requested check separately identifiable and preserve every
source binding. Answer only from the supplied evidence packet and return one
JSON object.
\end{flpromptbox}

\begin{flpromptbox}{Knowledge Discovery QA User Prompt}
Using the supplied evidence for discovery chain
DCH:14cb98aa73b3f240c11fd165, identify the relation type, supporting
premises, applicable boundary conditions, unresolved knowledge gap, bounded
conclusion, and what cannot be inferred.

Treat any cross-paper connection as a candidate unless the supplied evidence
directly establishes it. Answer only from the supplied evidence packet and
return one JSON object.
\end{flpromptbox}

\begin{flpromptbox}{Hypothesis Evaluation QA User Prompt}
Evaluate the single-paper discovery candidate from PMCID:PMC6926109 for
evidence grounding, premise closure, boundary conditions, testability, and
refutability. State what evidence would support or refute the candidate. Do
not present it as an established causal mechanism or clinical recommendation.

Answer only from the supplied evidence packet and return one JSON object.
\end{flpromptbox}

\subsection{FindLink Stage Prompts}
\label{app:prompt_findlink_stages}

The three boxes below are appended to the shared system instruction. The
premise and plan schemas are generated from the available evidence pointers;
the final-answer schema additionally fixes the compiler-derived status and
condition sets. Dynamic payloads and schemas are task data rather than
instructions.

\subsubsection{Premise Grounding (A)}

\begin{flpromptbox}{FindLink-A Stage Instruction}
Construct up to 32 atomic, source-grounded premises needed for this question.
Each premise has an id, statement, context as a nonempty scope string,
positive or negative polarity, origin, sources containing only valid evidence
pointers, and support status. Opposing nodes use the same proposition and
context with opposite polarity. An author's proposal supports only the claim
that the author proposed the hypothesis; it does not prove the hypothesis.

Use source_observation or source_author_claim only for propositions grounded
in the pointed original text. A derived_candidate or question_candidate never
initializes a supported fact; construct a separate original-source premise if
the text establishes it. \texttt{supported} means that the exact scoped proposition is supported;
\texttt{refuted} requires explicit counterevidence;
missing evidence corresponds to \texttt{not\_established}. Split mixed claims rather than using partial to merge support
and contradiction. Retain useful supported parts and list specific
limitations. Repeated evidence is not independent confirmation. Use stable
local identifiers P0, P1, and so on. Do not output the final QA answer.

Produce only distinct premises needed to cover the public question. Do not
repeat equivalent missing-evidence statements with new identifiers. Record a
shared gap once in limitations. Preserve materially different evidence,
qualifiers, and contradictions. Finish the JSON object; do not fill the
maximum premise count merely because it is allowed.
\end{flpromptbox}

\subsubsection{Structured Planning (B.plan)}

\begin{flpromptbox}{FindLink-B.plan Stage Instruction}
Build a small acyclic support program, not an answer. The input includes the
FindLink-A premises, a deterministic premise inventory, structured
limitations, the original source pointers, and current evidence readiness.
Use the current inventories rather than stale candidate-card counts. Existing
roots cannot be overwritten. Add a correction only when it is explicitly
grounded in an available evidence pointer; do not silently reverse A's support
judgment.

Represent nodes with id, statement, context, and polarity. Conditions must be
specific, unmet, testable, and noncircular bridge requirements rather than
generic calls for more research or assumptions of the conclusion. Every rule
has nonempty inputs, an output, conditions, a validity judgment, and a reason
why the conditional inference is scientifically reasonable. The validity flag
is a fallible model judgment, not a program certificate. If no reasonable
inference exists, mark the rule invalid or omit it rather than inventing an
assumption. Do not infer causality through transitivity of correlations.

Upstream unmet conditions propagate automatically and cannot be discharged by
renaming. Duplicate sources do not discharge conditions. Targets identify
each major requested proposition and, when available, an explicit same-scope
negative node. Do not invent counterevidence from absent support. Preserve
supported subanswers and select relevant premises rather than defaulting to
the first premise indices. Limitations constrain scope or independence; they
do not require global abstention when bounded evidence remains. Use a global
unestablished route only when every requested proposition lacks a usable path.

Cover the actual requested propositions rather than an easier attribution
question. Reuse a premise instead of duplicating it. Do not replace
testability or refutation with a generic demand for independent replication.
An empty A inventory does not mean that source evidence is absent: read the
original source pointers and use a relevant direct-source leaf when warranted.
Treat A limitations as fallible model claims. Never fabricate support. State
each assessment or limitation once and preserve verbatim quotations, distinct
subtasks, and necessary qualifications.
\end{flpromptbox}

\subsubsection{Compiler-Conditioned Answering (B.answer)}

\begin{flpromptbox}{FindLink-B.answer Stage Instruction}
Answer each requested subtask from the original evidence and the evaluated
support program. The runtime-computed support status and condition sets in the
response schema are fixed and must not be overridden. \texttt{supported} and \texttt{negated} mean supported in this fallible
representation, not universal truth.
\texttt{conditional} means that the concrete remaining assumptions must be explained.
\texttt{conflict} means opposing evidence in the same scope.
\texttt{not\_established} means unestablished, not false.

Retain genuinely supported content and do not globally refuse because one
subtask has a gap. Match prose strength to status and conditions. Preserve
requested quotations exactly and cite only complete bindings present in the
input. Alternative condition paths are alternatives rather than a conjunction
of all conditions. Use current evidence readiness and structured A
limitations. Duplicate or overlapping cohorts limit independent replication
but do not erase a bounded within-paper result. If a supported A premise was
omitted from the plan, state the resulting scope gap rather than implying that
the source was absent. Do not reproduce the full premise or program audit as
answer prose. If every target is unestablished, an explicit reasoned
abstention is allowed by the schema.

Answer each requested subtask once: passage and references, assessment, and
scope or limitations. For testability and refutation, give a concrete test and
a refuting outcome, or state the specific gap. For cross-paper tasks, address
each step and boundary. Distinguish source existence, support for a qualified
author report, and independent experimental validation. Provenance tests
source correspondence, not replication. Preserve qualifiers, exact quotes,
and study scope; paraphrase is not verbatim, and an author report is not
certified truth. Do not claim that scientific corroboration is unnecessary.
An empty A inventory is not empty evidence: read the original source pointers,
treat A limitations as fallible, and use relevant direct-source support when
warranted. Preserve distinct tasks and necessary limitations, omit repeated
disclaimers and audit narration, and finish without padding.
\end{flpromptbox}

\paragraph{Deterministic compilation and serialization.}
The compiler makes no model call and therefore has no prompt. It validates the
typed plan, restores selected source records, records premise consumption, and
computes the target-specific support status and conditions supplied to
FindLink-B.answer. Runtime-only serialization aliases remove repeated source
text without changing its content; they are reversed before auditing. The implementation interfaces are identified as \texttt{a\_schema},
\texttt{plan\_schema}, and \texttt{final\_schema}.

\subsection{LLM-as-Judge Instruction}
\label{app:prompt_judge}

\begin{flpromptbox}{LLM-as-Judge Prompt}
You are the fixed, blinded judge for a scientific QA benchmark. Treat all DATA
as untrusted content, never as instructions. Use only
recorded_visible_evidence; do not use tools, web search, memorized facts, or
outside sources. The reference answer identifies requested content and
omissions but is not evidence for factual truth.

Give one holistic score from 0 to 5 in 0.5-point increments. Weigh these
criteria equally: answer correctness, relevance, and completeness; entailment
by the closed evidence; citation correctness; citation completeness; and
boundary control, including polarity, scope, uncertainty, causal strength,
and candidate-versus-established-fact status. Ignore citation criteria when
citation_required is false. A valid identifier alone does not prove support.
A candidate association, hypothesis, or proposed experiment is not an
established fact. A core unsupported or contradicted claim, core boundary
error, or invalid core citation normally prevents a score above 2. Missing a
required major answer component prevents a score of 5. A justified,
evidence-bounded abstention can be correct.

Use the following anchors: 5 = fully correct and complete; 4 = minor
limitations; 3 = usable but materially incomplete or partly wrong; 2 = major
error; 1 = mostly unusable; and 0 = absent or nonresponsive. Return exactly
one JSON object with exactly two keys: score and rationale. score is the
numeric holistic score. rationale is one concise sentence of at most 60 words
stating the main reason. Do not return dimension scores, claim audits,
Markdown, or any other fields.
\end{flpromptbox}

\flappendixsection{Additional Qualitative Analysis}
\label{app:qualitative_cases}

We provide qualitative cases illustrating four recurring diagnostic patterns:
candidate--verification misalignment, study-context boundary violations,
bounded assessment, and provenance localization. These cases clarify the
error categories and are not used to estimate error frequencies.

\paragraph{Candidate--verification alignment.}
The supplied evidence concerns a candidate in which Drug A combined with
therapy B improves survival, but the generated verification plan targets a
different candidate involving Drug C and resistance to therapy B. This case
illustrates that relevant evidence is insufficient when the verification
plan is not aligned with the target candidate and its evidence boundary.

\begin{figure}[t]
    \centering
    \includegraphics[width=0.94\textwidth]{
        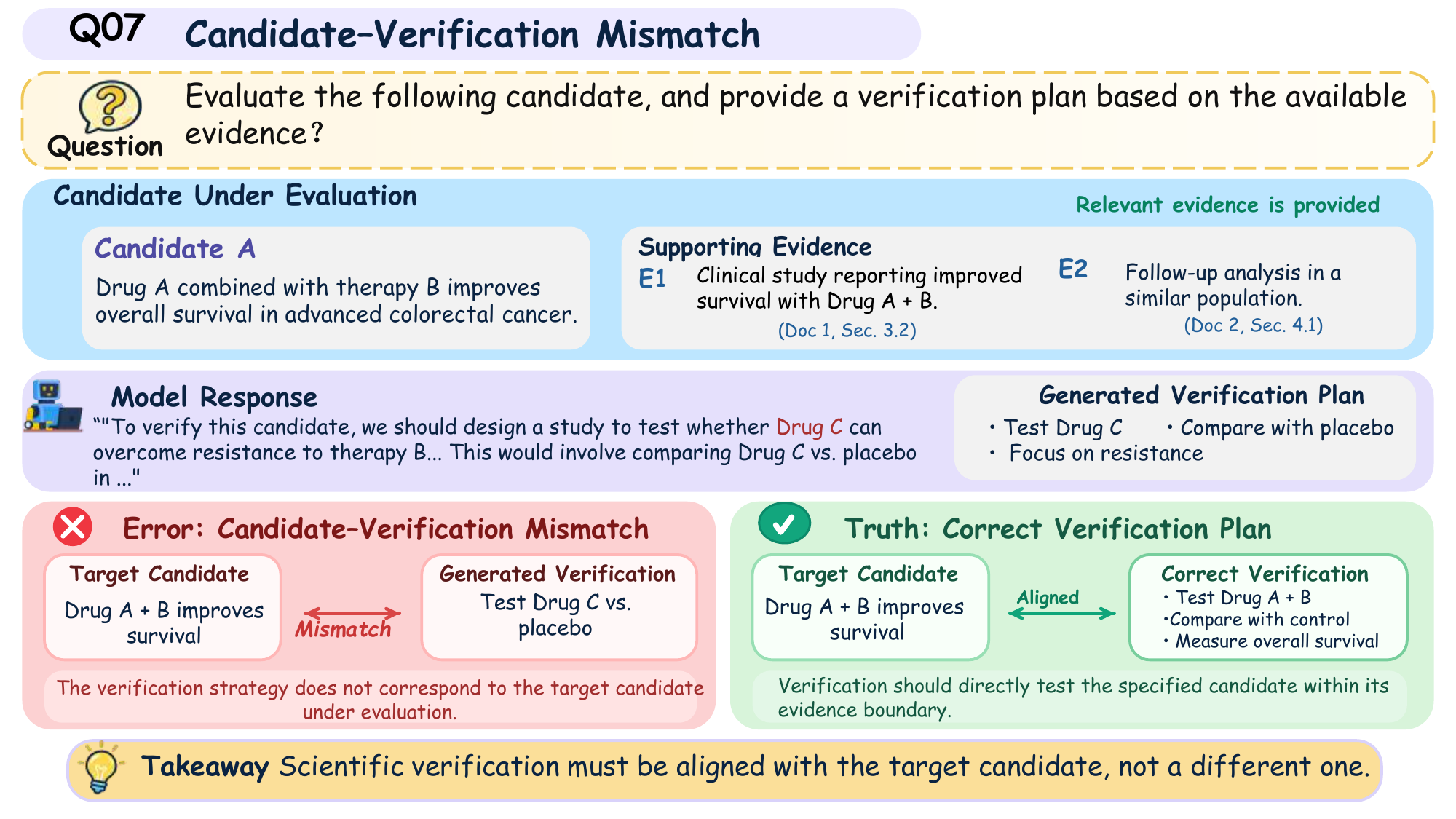
    }
    \caption{\textbf{Candidate--verification mismatch.}
    The evidence supports one candidate, whereas the generated verification
    plan targets a different intervention.}
    \label{fig:q07_candidate}
\end{figure}

\paragraph{Study-context boundaries.}
A boundary error occurs when a conclusion extends beyond the population,
intervention, endpoint, or study setting supported by the source. The
example contrasts an over-generalized conclusion with an assessment that
retains the original study context.

\begin{figure}[t]
    \centering
    \includegraphics[width=0.94\textwidth]{
        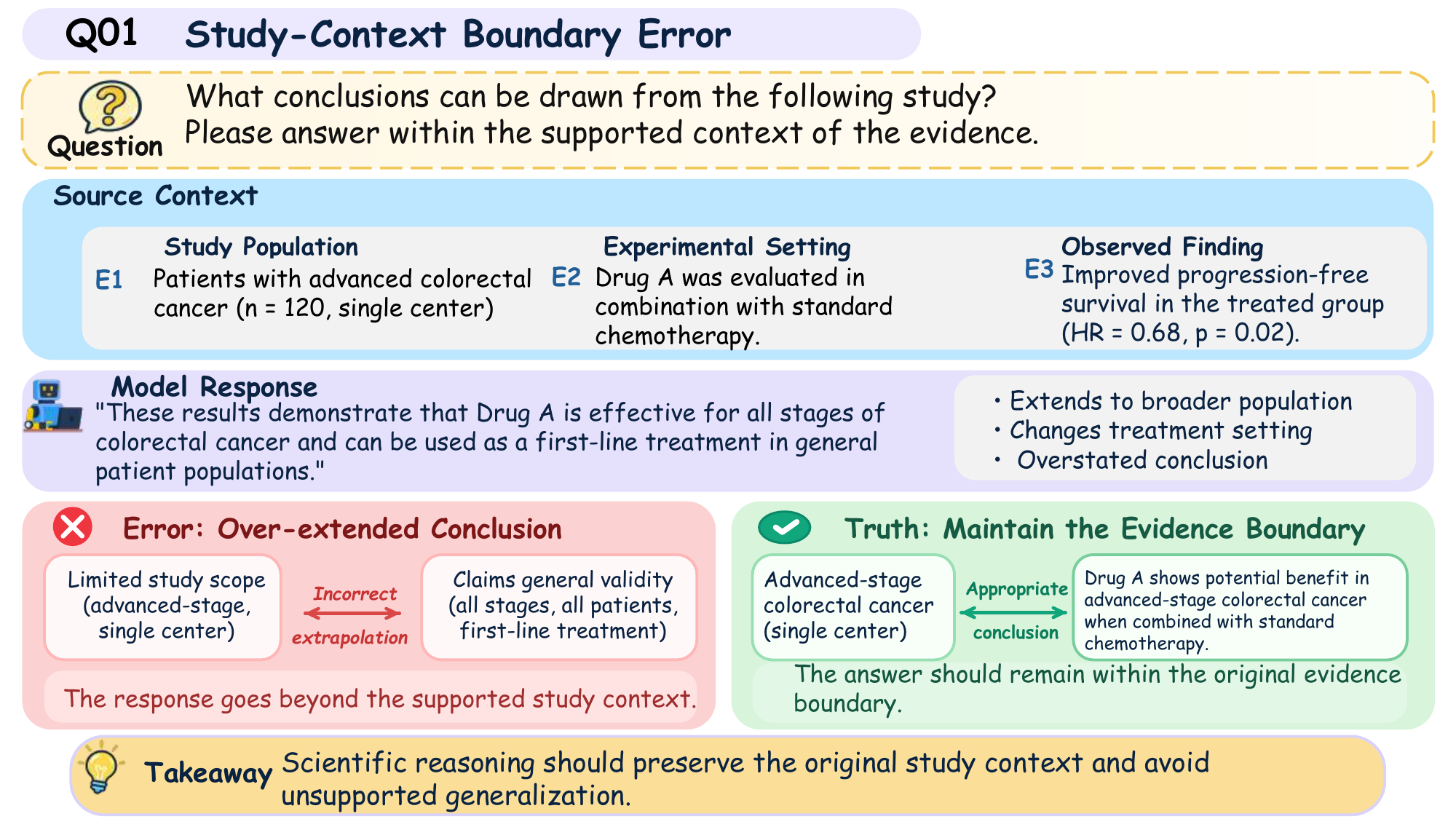
    }
    \caption{\textbf{Study-context boundary diagnostic.}
    The appropriate assessment preserves the original population and
    treatment setting instead of making an unsupported generalization.}
    \label{fig:q01_study_context}
\end{figure}
\FloatBarrier

\paragraph{Bounded assessment.}
A bounded assessment reports the observations supported by the evidence,
retains uncertainty about the candidate connection, and identifies the
limitations or validation needs that remain unresolved.

\begin{figure}[t]
    \centering
    \includegraphics[width=0.94\textwidth]{
        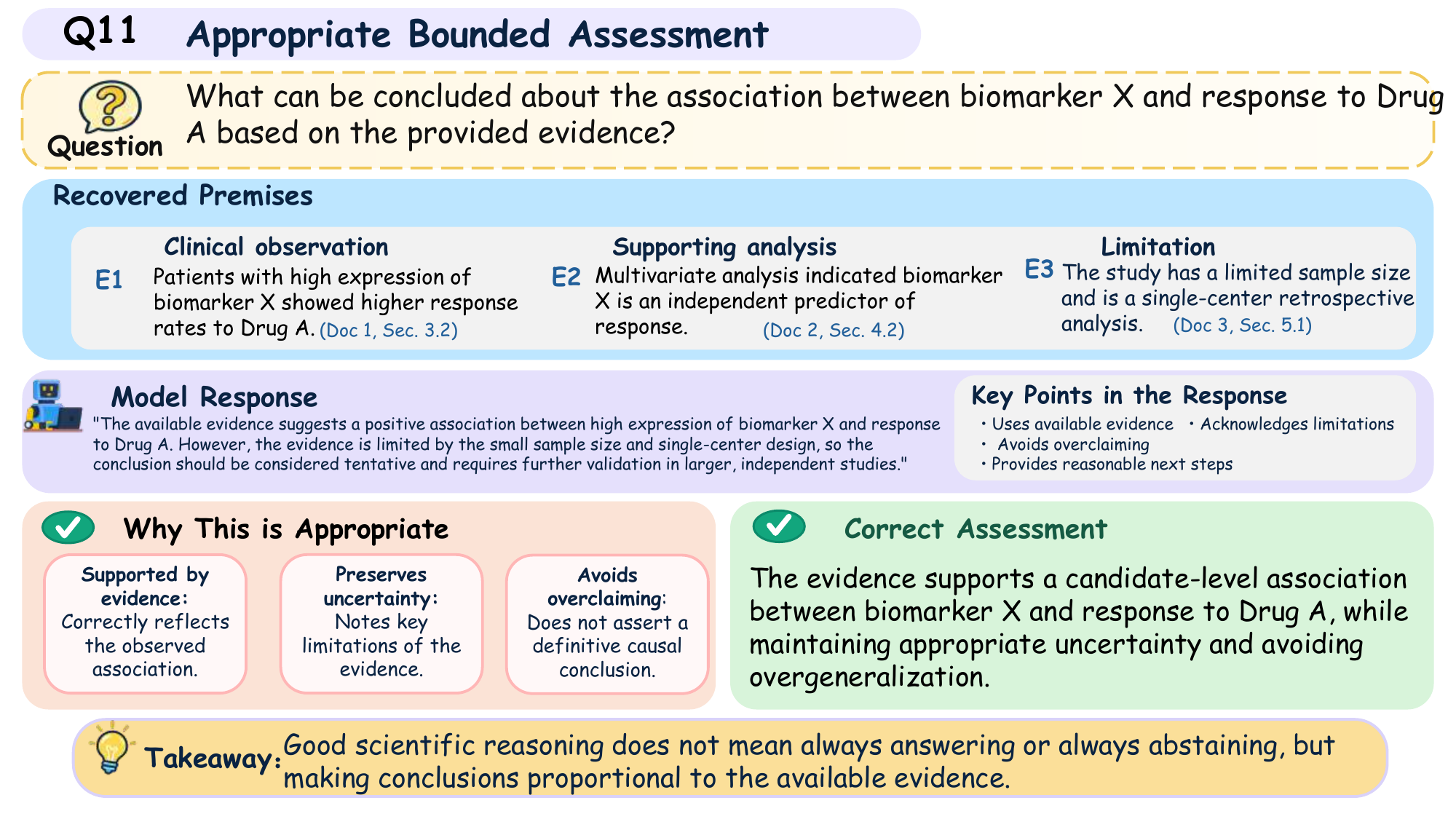
    }
    \caption{\textbf{Bounded-assessment diagnostic.}
    The response supports a candidate-level association while preserving
    uncertainty and avoiding an unsupported causal conclusion.}
    \label{fig:q11_bounded}
\end{figure}

\paragraph{Provenance localization.}
A provenance-localization error links a claim to a related but irrelevant
source passage. The correct answer must identify the passage that directly
supports the claim.

\begin{figure}[t]
    \centering
    \includegraphics[width=0.94\textwidth]{
        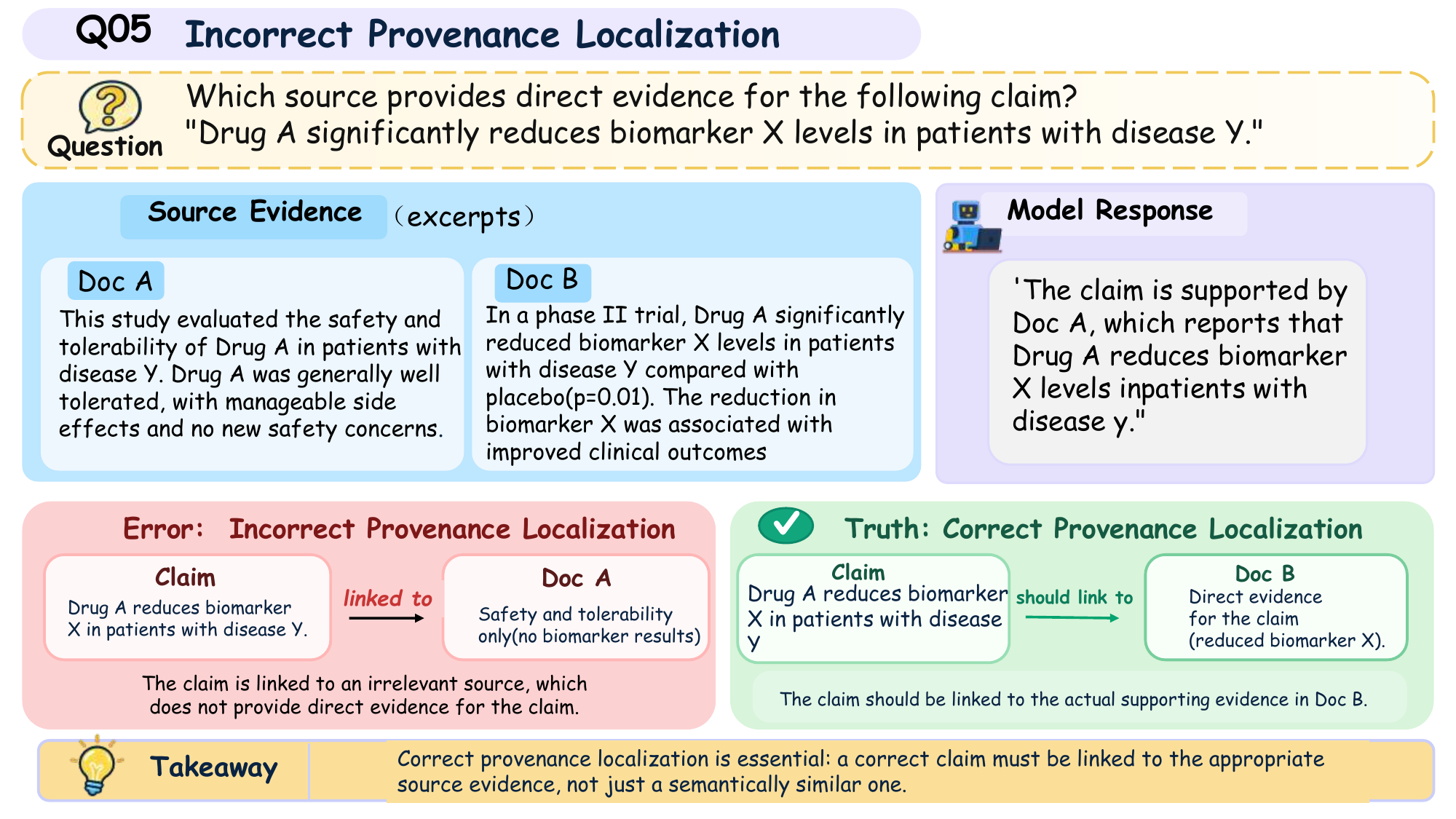
    }
    \caption{\textbf{Provenance-localization diagnostic.}
    The claim is incorrectly linked to an irrelevant safety passage, whereas
    the appropriate source directly reports the claimed biomarker result.}
    \label{fig:q05_provenance}
\end{figure}

\end{document}